\documentclass{article} %

\usepackage[normalem]{ulem}

\usepackage{natbib}
\usepackage{authblk}
\usepackage{times}
\usepackage[scaled=0.999]{helvet}
\usepackage{microtype}

\usepackage[letterpaper, left=1.5in, right=1.5in, top=1in, bottom=1in]{geometry}

\makeatletter
\renewenvironment{abstract}{%
  \if@twocolumn\section*{\abstractname}\else
    \small
    \begin{center}
      {\bfseries\abstractname\vspace{-.5em}\vspace{\z@}}
    \end{center}
    \list{}{\leftmargin 3pc\relax\rightmargin\leftmargin}
    \item\relax
  \fi}
  {\if@twocolumn\else\endlist\fi}
\makeatother

\skip\footins 9pt plus 4pt minus 2pt
\def\footnoterule{\kern-3pt \hrule width 12pc \kern 2.6pt }

\leftmargini\leftmargin \leftmarginii 2em
\belowdisplayskip \abovedisplayskip
\makeatletter
\def\normalsize{\@setsize\normalsize{11pt}\xpt\@xpt}
\def\small{\@setsize\small{10pt}\ixpt\@ixpt}
\def\footnotesize{\@setsize\footnotesize{10pt}\ixpt\@ixpt}
\def\scriptsize{\@setsize\scriptsize{8pt}\viipt\@viipt}
\def\tiny{\@setsize\tiny{7pt}\vipt\@vipt}
\def\large{\@setsize\large{14pt}\xiipt\@xiipt}
\def\Large{\@setsize\Large{16pt}\xivpt\@xivpt}
\def\LARGE{\@setsize\LARGE{20pt}\xviipt\@xviipt}
\def\huge{\@setsize\huge{23pt}\xxpt\@xxpt}
\def\Huge{\@setsize\Huge{28pt}\xxvpt\@xxvpt}
\normalsize
\makeatother

\usepackage{amsmath}
\usepackage{amsthm}
\usepackage{bm}
\usepackage{amssymb}

\usepackage{graphicx}
\usepackage{wrapfig}
\usepackage[svgnames]{xcolor}
\usepackage{subcaption}

\definecolor{dartmouthgreen}{HTML}{017d39}
\definecolor{linkblue}{HTML}{0022f0}
\definecolor{lightgray}{gray}{0.95}

\usepackage{array}
\usepackage{booktabs}
\usepackage{multirow}
\usepackage{makecell}
\usepackage{adjustbox}
\usepackage{siunitx}

\usepackage{algorithm}
\usepackage{algpseudocode}

\usepackage{subcaption}

\usepackage{enumitem}

\usepackage{url}
\usepackage{xcolor}
\usepackage[colorlinks=true, allcolors={blue!70!black}]{hyperref}

\theoremstyle{definition}

\newenvironment{compacttable}[1][]%
  {%
    \begingroup
    \captionsetup{aboveskip=2pt,belowskip=0pt}%
    \renewcommand{\arraystretch}{0.9}%
    \setlength{\tabcolsep}{3.5pt}%
    \footnotesize
    #1%
  }%
  {%
    \endgroup
  }

  {%
    \setlength{\topsep}{0pt}%
    \setlength{\parskip}{0pt}%
    \begin{center}%
  }%
  {%
    \end{center}%
  }

\title{\textbf{Adaptive Model Inversion Attacks Generalize a Privacy-Robustness Tradeoff}}

\author[1]{{Shailen Smith}}
\author[1]{{Rasmus Torp}}
\author[1,2]{{Adam Breuer}\thanks{\textsf{\textbf{{Breuer Lab gratefully acknowledges the support of the  \textls[120]{OpenAI}  \textsc{Cybersecurity Grant.}}}}}}

\affil[1]{Department of Computer Science, \textls[36]{Dartmouth College}}
\affil[2]{Department of Government, \textls[36]{Dartmouth College}}

\date{\vspace{-0.7em}\{shailen.k.smith.gr, rasmus.torp.gr, adam.breuer\}\textls[30]{@dartmouth.edu}}

\begin{document}

\maketitle

\begin{abstract}
In this paper, we show that standard evaluations of high-resolution Model Inversion Attacks (MIAs) significantly underestimate training-data privacy leakage. State-of-the-art privacy defenses, standard training techniques such as MixUp and Adversarial Training, and undefended models all leak training images at rates $1.16$ to $6.59\times$ higher on FaceScrub under simple adaptive changes to the attack, with the largest increases among defenses reporting the strongest privacy. We further show that measured leakage depends on the feature basis of the external classifier used to evaluate reconstructions: for the same reconstructed images, an adversarially trained Inception evaluator identifies the targeted identity at different rates than the standard Inception evaluator. Our results suggest that standard MIA evaluation can mistake optimization and measurement failures for privacy.

These underestimated leakage rates also concealed a broader relationship between privacy and adversarial robustness. Once we adapt the attack and vary the evaluator, reconstruction leakage closely tracks adversarial robustness across recent defenses and standard training regimes, suggesting that robustness provides an attack-agnostic proxy for reconstruction vulnerability that applies far more broadly than previously theorized. This raises an open question: can a practical defense reduce training-data reconstruction without paying a corresponding cost in adversarial robustness?
\end{abstract}

\section{Introduction}
\label{sec:introduction}
Model inversion attacks (MIAs) provide an empirical test of training-data privacy leakage under a powerful adversary. In the high-resolution vision setting, an attacker receives white-box access to a trained model, substantial auxiliary data, and a generative image prior, and attempts to reconstruct recognizable representatives of private training classes~\citep{fredrikson_model_2015,zhang_secret_2020,chen_knowledge-enriched_2021,struppek_plug_2022,an_mirror_2022,nguyen_re-thinking_2023,peng2024ppdg,qiu_closer_2024}. Successful reconstruction therefore establishes that recognizable information about the private training data, such as extractable facial geometry, was exposed by the model. MIA evaluation is therefore an empirical worst-case lower bound on training-data exposure: measured leakage is only as strong as the attacker used to measure it. A low training data reconstruction rate may indicate a private model. Alternatively, it may simply reflect a failed attack.

\textbf{The limited-scope privacy--robustness tradeoff and its outliers.}
Recent work proposed a surprisingly simple explanation for this leakage. \citet{torp2026reducing} found that, across three generic MIA defenses (MID, BiDO, and TL-DMI) \citep{wang_improving_2020, peng_bilateral_2022, ho_model_2024}, reconstruction leakage could be predicted almost perfectly from clean and adversarial accuracy (pseudo-$R^2=0.952$). They attribute this relationship to the model's feature basis: robust features are theorized to be semantically meaningful and therefore useful for visual reconstruction \cite[see, e.g.,][]{ilyas_adversarial_2019,engstrom_adversarial_2019,tsipras2018robustness}, whereas non-robust features remain predictive but are visually imperceptible. Deliberately shifting models toward non-robust features correspondingly reduced reconstruction. The result suggested a striking connection between two normally separate Trustworthy-ML objectives: robustness to adversarial examples and data privacy.

The result is a fascinating theory bridging robustness and privacy, yet it is also theorized \emph{not} to generalize to many of the defenses and training techniques that are relevant in practice. Specifically, the original result explicitly restricted the scope of the theorized privacy--robustness tradeoff to three generic defenses, MID, BiDO, and TL-DMI, and the theory explicitly excludes recent high-performance defenses such as NegLS, RoLSS, RoLSS-SSF, and Trap-MID~\citep{struppek_be_2024,koh_vulnerability_2024,liu2024trapmid} that interfere with the gradients used by contemporary MIAs. Also excluded from the theory are standard data augmentation and training techniques such as MixUp and Adversarial Training (AT)~\citep{zhang2018mixup,madry_towards_2017} that practitioners commonly deploy to improve generalization. 

\begin{wrapfigure}{r}{0.43\textwidth}
    \centering
    \vspace{-12pt}
    \includegraphics[width=\linewidth]{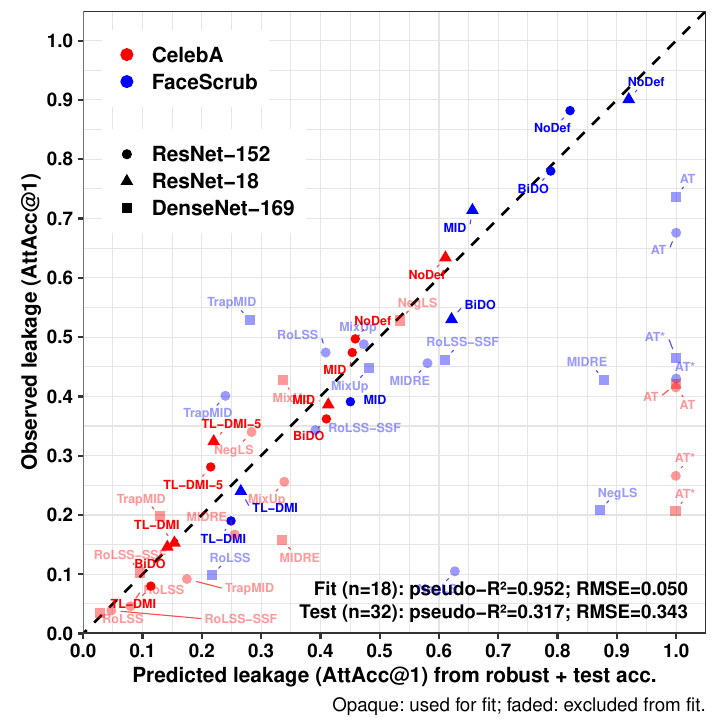}
    \caption{The privacy--robustness relationship in \citet{torp2026reducing} does not generalize under standard MIA evaluation. Opaque points were included in the original \citet{torp2026reducing} regression.}
    \label{fig:regression-outlier}
    \vspace{-20pt}
\end{wrapfigure}

Fig.~\ref{fig:regression-outlier} applies the published regression coefficients to this broader set. The fit collapses from pseudo-$R^2=0.952$ (RMSE $=0.050$) on the original defenses to pseudo-$R^2=0.317$ (RMSE $=0.343$) on this out-of-sample set. Most strikingly, AT models leak far less than their extreme robustness predicts, while recent NegLS and MIDRE~\citep{tran2025random} defenses appear to obtain substantial privacy without the corresponding robustness cost. The AT result is particularly surprising because adversarial training produces the robust feature basis implicated by the theory and has previously been shown to increase vulnerability to model inversion~\citep{mejia_robust_2019}.

There are two possible explanations. The first is simply that adversarial robustness was a strong correlate of privacy for three particular defense families, but as \citet{torp2026reducing} theorized, it is not a general property of model inversion.

\textbf{Are defenses' privacy claims overfit to an evaluation that hides generalizable properties?}\\In this paper, we consider a second possibility: otherwise generalizable privacy properties that apply across defenses and training techniques in the field are spuriously obscured because contemporary defense evaluations are overfit to a fixed configuration of attacks and evaluations, resulting in falsely inflated claims of privacy improvements among top defenses. The problem we describe is familiar from adversarial robustness. There, defenses against adversarial examples repeatedly appeared robust because they were evaluated against attacks whose optimization failed on the defense, motivating adaptive attacks that  account for the defense being evaluated~\citep{tramer2020adaptive,athalye2018obfuscated}. The analogous principle has not become standard in MIA. Recent hi-res MIA defenses are often evaluated using fixed attacks such as  PPA \citep{struppek_plug_2022} with the attack's loss, optimizer, and hyperparameters set to (non-adaptive) published values, even when the defense itself changes the gradients on which that optimization depends (see App. \ref{app:related-work-adaptive-mia}). %

A recent geometric account of model inversion suggests a mechanism by which the privacy--robustness relationship could generalize beyond the defenses on which it was originally fit. \citet{peng2026generative} observe that generative MIAs do not follow the full input-space gradient: because optimization proceeds through a generator $G(z)$, the attack can move only along directions in the tangent space of the generator manifold. This projection suppresses off-manifold components of the inversion gradient, and Peng et al.\ show that inversion succeeds more readily when the target model's loss gradients are better aligned with this manifold. Relatedly, \citet{srinivas2023models} tie perceptually aligned input gradients closely to adversarial robustness: for Bayes-optimal classifiers, perceptual alignment with the signal manifold corresponds to off-manifold robustness, while empirically several forms of robust training induce off-manifold robustness, and in turn, perceptually aligned gradients. These are not the same notions of alignment: Peng et al.\ measure alignment to the generator manifold during inversion, whereas Srinivas et al.\ study alignment to the signal manifold of the  data distribution. However, they suggest a common geometric possibility. Models whose predictive gradients concentrate on semantically meaningful, on-manifold directions may be both more adversarially robust and easier for a generator-constrained attack to reconstruct. If so, defenses that appear to escape the privacy--robustness relationship because a fixed attack fails to optimize may not remove the underlying exposure of their training data; they may instead make training data features harder for a particular attack optimizer configuration to extract.

\textbf{Our contribution.}
We test these apparent exceptions under adaptive model inversion across 13 defenses and generic training techniques, spanning nine MIA defenses, an undefended baseline, MixUp, and two Adversarial Training (AT) strengths; two standard high-resolution target architectures (ResNet-152 and DenseNet-169); and the two main high-resolution MIA datasets (FaceScrub and CelebA), under PPA and IF-GMI. We make simple, generic changes to the attack loss function and optimization while holding the attack generator and target model fixed.

We show that under these simple optimization changes, SOTA defenses' measured privacy leakage increases by factors of $1.16$--$6.59\times$ in terms of standard AttAcc@1 on FaceScrub, with larger differences between vanilla and adaptive attacks corresponding to several of the recent defenses with the highest claimed performance. On ResNet-152, an adaptive variant increases AttAcc@1 for 25 of 26 model--dataset pairs. For example, NegLS AttAcc@1 on FaceScrub changes from $0.105$ to $0.692$ and TL-DMI from $0.151$ to $0.459$; undefended models and models trained with MixUp or AT leak $1.18$--$2.61\times$ more with our techniques. We also show that the leakage increases generalize to, e.g., CelebA data, DenseNet-169 architectures, and our techniques also apply to IF-GMI attacks.

Attack optimization is only one source of measurement dependence. High-resolution MIAs conventionally score whether a reconstruction recovered its target identity using an independent Inception-v3 classifier~\citep{struppek_plug_2022, szegedy_rethinking_2016}. But the feature account of \citet{torp2026reducing} suggests that this measurement may depend on the evaluator's own feature basis. We therefore score the same reconstructions with both standard and adversarially trained Inception evaluators. Neither evaluator is treated as ground truth; disagreement between them measures how strongly reported leakage depends on the evaluator's feature basis. For example, a generic adversarially-trained RN-152 model's AttAcc@1 leakage changes from $0.430$ under the standard evaluator to $0.901$ under the robust evaluator, on the same reconstructed images. The results indicate that evaluator models used to compute
the main privacy metric in the high-resolution literature underestimate leakage, particularly where the target model relies on robust, visually meaningful features.

Finally, we return to the regression in Fig.~\ref{fig:regression-outlier}. Combining adaptive attacks with the robust evaluator yields pseudo-$R^2=0.740$ (RMSE $=0.156$), without adding the held-out defenses or training regimes to the fitting set. This suggests that the apparent failure of the tradeoff is partly a property of the standard evaluation choices rather than evidence that recent defenses break the relationship: once the attack is adapted, the same robustness predictors generalize substantially better to defenses and training regimes that were never used to fit them, including defenses explicitly designated as exceptions in \citet{torp2026reducing}. The  fit further improves under a robust evaluator.

\textbf{Implications for a generic privacy--robustness tradeoff.} Our results have two implications for the MIA and Trustworthy-AI fields. First,  our results suggest a broader interpretation of the privacy--robustness relationship. Adversarial robustness is not merely a correlate that happened to fit three MIA defenses. In the high-resolution image-classification setting we study, it captures a property of the learned representation that is also implicated by recent geometric accounts of inversion: whether semantically meaningful directions are available to be reconstructed. Gradient manipulation can make those directions harder for a certain optimizer to find, but does not necessarily remove them.

This frames a key question for future research: is it possible to design a defense against training data reconstruction that obtains privacy without a corresponding sacrifice of adversarial robustness?

\textbf{Implications for MIA privacy defense evaluation.} Second, MIA evaluations have substantially underestimated the degree to which state-of-the-art privacy defenses leak information about their training data. By developing defenses that change the gradients on which attack optimization depends, while continuing to evaluate them with fixed attacks, fixed hyperparameters, and a single Inception feature basis, the field has recreated a problem familiar from adversarial examples: apparent privacy can reflect failure of the evaluation procedure rather than removal of the underlying vulnerability. We observe both forms of dependence here; simple adaptive changes recover substantially more leakage from most defenses, while the same reconstructed images can receive sharply different privacy scores under an alternative evaluator feature basis.

This changes the appropriate standard for evaluating whether an empirical MIA defense succeeds.  A privacy defense should survive an attacker that adapts to the defense, and its reconstructed images should be evaluated by more than one arbitrarily chosen classifier feature basis. Otherwise, we risk calling an optimization failure `privacy'.

\section{Adaptive Attacks and Robust Leakage Evaluation under MIA}
\label{sec:methods}
Our goal is to test whether the measured privacy gains of model inversion defenses persist when we change two fixed choices in the standard evaluation: how the attack optimizes reconstructions and how an external classifier scores them. We keep the trained target models fixed throughout.

\subsection{Adaptive attacks}
\label{sec:adaptive-ppa}
A defense may make reconstruction difficult across attack settings. Alternatively, it may change the attacker's loss landscape in a way that frustrates the particular loss and learning rate used in standard evaluations. We test whether the reported privacy gains survive two ordinary changes to the white-box Plug-and-Play Attack (PPA) of \citet{struppek_plug_2022} used as the primary high-resolution evaluation attack in recent MIA defense evaluations \citep[e.g.,][]{torp2026reducing, tran2025random, liu2024trapmid, struppek_be_2024}. Specifically, we consider replacing its loss and increasing its learning rate by a factor of ten. The generator and target model remain fixed. The same changes can be applied to other attacks that optimize generator latents using target-model gradients.

\textbf{Attack losses.} Let the GAN used by generative model inversion attacks be denoted as $G$, and let the target classifier $T$ be trained on $C$ classes. PPA optimizes a StyleGAN-2 \citep{karras_analyzing_2020} latent vector $z$ for a target identity $c$ using gradients through the target classifier's logits $o=T(G(z))\in\mathbb{R}^{C}$. Standard PPA uses a Poincar\'e loss and an initial Adam learning rate of $0.005$, reduced after iterations 30 and 40. Poincar\'e loss was introduced to address a particular failure mode of cross-entropy: its gradients vanish as target confidence approaches one, which can prevent late-stage refinement of poorly initialized latents \citep{struppek_plug_2022}. The defenses we study can create different optimization failures. Some alter the loss landscape such that PPA encounters local modes before reaching high target confidence; others, such as NegLS, directly suppress the gradients used by the attack. We therefore test whether PPA's fixed objective is itself responsible for part of the claimed privacy gain by replacing Poincar\'e with either vanilla cross-entropy or cross-entropy with negative label smoothing (an attack adaptation inspired by the NegLS defense). Let $p_j(o)=\exp(o_j)/\sum_k\exp(o_k)$, and denote the one-hot target by $e_c$. 

Then, both alternatives take the form
\begin{equation}
    q^{(c)}=(1-\alpha)e_c+\frac{\alpha}{C}\mathbf{1},\qquad
    \mathcal{L}_{\alpha}(o,c)=-\sum_{j=1}^{C}q^{(c)}_j\log p_j(o),
    \label{eq:adaptive-loss}
\end{equation}
with $\alpha=0$ for ordinary cross-entropy and $\alpha=-0.05$ for negative label smoothing. With $\alpha<0$, the off-target weights are negative and the gradient with respect to the target logit does not vanish as $p_c(o)$ approaches one. Negative label smoothing therefore retains a class-directed gradient even in the high-confidence regime that originally motivated PPA's substitution of Poincar\'e loss. We apply both losses to all target models, including those trained without negative label smoothing.

\textbf{Tenfold learning rate.} A defense that introduces local modes into PPA's optimization landscape may impede the default optimizer's progress without preventing reconstruction at some other step size. We test this possibility in the simplest conceivable way by raising the initial learning rate tenfold from $0.005$ to $0.05$ and reducing it after iterations $20$, $30$, and $40$. We refer to our cross-entropy and negatively smoothed attacks at this tenfold-increased learning rate as \emph{C-LR} and \emph{N-LR}. %

\subsection{Robust evaluation}
\label{sec:paired-evaluation}
Separately, a defense's measured privacy gain may reflect how its target model's feature basis aligns with the evaluator's, rather than how much recognizable facial information the attack recovers. The standard Attack Accuracy (AttAcc@1) metric uses an independent Inception-v3 \citep{szegedy_rethinking_2016} classifier as a proxy for whether a reconstructed face is recognizable as the targeted identity \citep{struppek_plug_2022}. That proxy depends on the features Inception itself uses to recognize a face. As \citet{torp2026reducing} show, defenses can shift the target model toward non-robust identity features. A generator may encode these predictive but semantically meaningless features in its reconstructions, and a conventionally trained Inception model may classify from them, even if the reconstructed face is not recognizable to a person. The reverse is also possible: Inception may miss visible facial structure recovered from a target that relies more heavily on robust features. 

Accordingly, we test whether defenses' privacy gains persist under an evaluator trained to place more weight on robust features associated with visible facial structure. For each dataset, we score the \emph{same saved reconstructions} using the standard Inception-v3 evaluator and an independently adversarially trained Inception-v3 identity classifier. AttAcc@1 and Robust-AttAcc@1 are the respective fractions for which each evaluator's top prediction matches the attacked identity. We report both scores and each evaluator's accuracy on held-out natural images. The attack, target model, and reconstruction remain fixed in each comparison. Thus, disagreement between AttAcc@1 and Robust-AttAcc@1 empirically lower-bounds the degree to which measured defense performance depends on a single arbitrarily chosen classifier feature basis.

\section{Experiments}
\label{sec:experiments}
Our goal in this section is to test whether the reported privacy gains of current model inversion defenses survive our adaptive attack techniques across the standard high-resolution evaluation settings in the defense literature. To accomplish this, we run unmodified SOTA PPA attacks alongside PPA attacks adapted with our two natural substitute loss functions with a tenfold increased learning rate, C-LR and N-LR. App. \ref{deferred-model-inversion-results} also applies our techniques to alternate IF-GMI attacks \citep{qiu_closer_2024}.

\textbf{Experimental scope.} We consider 13 target-model conditions: 9 SOTA MIA baselines, 3 models trained using standard techniques for improving generalization (MixUp and AT under two perturbation magnitudes $\epsilon$), and an undefended model (NoDef). The 9 baselines include the generic information-dependency defenses used to estimate the original privacy--robustness relationship (MID \citep{wang_improving_2020}, BiDO \citep{peng_bilateral_2022}, and TL-DMI \citep{ho_model_2024}),  recent high-performance and anti-gradient defenses (NegLS \citep{struppek_be_2024}, Trap-MID \citep{liu2024trapmid}, RoLSS \citep{koh_vulnerability_2024}, RoLSS-SSF,  MIDRE \citep{tran2025random}), and a causal instrument (AT-AT \citep{torp2026reducing}).  We evaluate defenses on the two main hi-res MIA datasets (FaceScrub \citep{ng_data-driven_2014} and CelebA \citep{liu_deep_2015}), across the two main hi-res architectures (ResNet-152 \citep{he_deep_2015} and DenseNet-169 \citep{huang_densely_2017}), under  PPA and our C-LR, and N-LR techniques. This yields $13$ target-models $\times$ $3$ PPA variants $\times$ $2$ datasets $\times$ $2$ architectures $=156$ conditions. App. \ref{deferred-model-inversion-results}  shows our techniques applied to IF-GMI attacks \citep{qiu_closer_2024}. 

\textbf{Standard \& robust privacy metrics.} For each defense, we report standard test accuracy and 5 reconstruction metrics. We report AttAcc@1 and AttAcc@5, the fractions for which an independent Inception-v3 evaluator ranks the target class first or among its top five predictions, respectively. We also report Robust-AttAcc@1 and Robust-AttAcc@5, which are the same metrics but computed with an adversarially-trained Inception-v3 model. The gap between AttAcc@1 and Robust-AttAcc@1 lower-bounds the sensitivity of measured defense performance to the evaluator's feature basis. We also report the L2-Face metric, which captures reconstructions' distance from private target images in FaceNet representation space.
\begin{table}[t]
\newcommand{\ppabar}{\rule[0.4ex]{0.80em}{0.23em}}
\centering
\small
\setlength{\tabcolsep}{3pt}
\renewcommand{\arraystretch}{1.05}
\caption{ResNet-152 adaptive attack results. Values show standard
PPA $\rightarrow$ worst-case over \{PPA, PPA N-LR,
and PPA C-LR\}, computed
independently. Markers in ($\cdot$) denote the strongest attack. }
\label{tab:resnet152-ppa-worst-case}
\begin{adjustbox}{max width=\linewidth}
\begin{tabular}{@{}lccc@{\hspace{12pt}}ccc@{}}
\toprule
& \multicolumn{3}{c}{\textbf{FaceScrub}} & \multicolumn{3}{c}{\textbf{CelebA}} \\
\cmidrule(lr){2-4}\cmidrule(lr){5-7}
\textbf{Model} & \textbf{TestAcc} $\uparrow$ & \textbf{AttAcc@1} $\downarrow$ & \textbf{L2-Face} $\uparrow$ &
\textbf{TestAcc} $\uparrow$ & \textbf{AttAcc@1} $\downarrow$ & \textbf{L2-Face} $\uparrow$ \\
\midrule
NoDef & 0.943 & $0.823 \rightarrow 0.969\,{\scriptstyle (\bigstar)}$ & $0.806 \rightarrow 0.712\,{\scriptstyle (\bigstar)}$ & 0.845 & $0.459 \rightarrow 0.695\,{\scriptstyle (\bigstar)}$ & $0.852 \rightarrow 0.838\,{\scriptstyle (\bigstar)}$ \\
\midrule
MID & 0.950 & $0.391 \rightarrow 0.590\,{\scriptstyle (\blacktriangle)}$ & $0.950 \rightarrow 0.867\,{\scriptstyle (\blacktriangle)}$ & 0.802 & $0.474 \rightarrow 0.781\,{\scriptstyle (\bigstar)}$ & $0.821 \rightarrow 0.709\,{\scriptstyle (\bigstar)}$ \\
BiDO & 0.920 & $0.754 \rightarrow 0.876\,{\scriptstyle (\bigstar)}$ & $0.832 \rightarrow 0.797\,{\scriptstyle (\bigstar)}$ & 0.747 & $0.362 \rightarrow 0.521\,{\scriptstyle (\bigstar)}$ & $0.923 \rightarrow 0.876\,{\scriptstyle (\bigstar)}$ \\
TL\mbox{-}DMI & 0.891 & $0.151 \rightarrow 0.459\,{\scriptstyle (\bigstar)}$ & $1.088 \rightarrow 0.954\,{\scriptstyle (\bigstar)}$ & 0.814 & $0.281 \rightarrow 0.513\,{\scriptstyle (\bigstar)}$ & $0.929 \rightarrow 0.863\,{\scriptstyle (\bigstar)}$ \\
RoLSS & 0.886 & $0.474 \rightarrow 0.631\,{\scriptstyle (\bigstar)}$ & $0.858 \rightarrow 0.832\,{\scriptstyle (\bigstar)}$ & 0.537 & $0.046 \rightarrow 0.072\,{\scriptstyle (\bigstar)}$ & $1.147 \rightarrow 1.117\,{\scriptstyle (\blacktriangle)}$ \\
RoLSS\mbox{-}SSF & 0.869 & $0.343 \rightarrow 0.492\,{\scriptstyle (\bigstar)}$ & $0.918 \rightarrow 0.894\,{\scriptstyle (\bigstar)}$ & 0.428 & $0.039 \rightarrow 0.045\,{\scriptstyle (\blacktriangle)}$ & $1.183 \rightarrow 1.183\,{\scriptstyle (\text{\ppabar})}$ \\
MIDRE & 0.941 & $0.456 \rightarrow 0.704\,{\scriptstyle (\bigstar)}$ & $0.891 \rightarrow 0.815\,{\scriptstyle (\bigstar)}$ & 0.756 & $0.167 \rightarrow 0.258\,{\scriptstyle (\bigstar)}$ & $0.987 \rightarrow 0.951\,{\scriptstyle (\bigstar)}$ \\
Trap\mbox{-}MID & 0.926 & $0.401 \rightarrow 0.873\,{\scriptstyle (\bigstar)}$ & $0.980 \rightarrow 0.777\,{\scriptstyle (\bigstar)}$ & 0.806 & $0.092 \rightarrow 0.302\,{\scriptstyle (\bigstar)}$ & $1.358 \rightarrow 1.150\,{\scriptstyle (\bigstar)}$ \\
NegLS & 0.911 & $0.105 \rightarrow 0.692\,{\scriptstyle (\bigstar)}$ & $1.217 \rightarrow 0.796\,{\scriptstyle (\bigstar)}$ & 0.805 & $0.340 \rightarrow 0.359\,{\scriptstyle (\bigstar)}$ & $0.908 \rightarrow 0.897\,{\scriptstyle (\bigstar)}$ \\
\midrule
AT\mbox{-}AT & 0.941 & $0.058 \rightarrow 0.058\,{\scriptstyle (\text{\ppabar})}$ & $1.225 \rightarrow 1.225\,{\scriptstyle (\text{\ppabar})}$ & 0.819 & $0.049 \rightarrow 0.055\,{\scriptstyle (\bigstar)}$ & $1.245 \rightarrow 1.245\,{\scriptstyle (\text{\ppabar})}$ \\
\midrule
MixUp & 0.971 & $0.488 \rightarrow 0.906\,{\scriptstyle (\blacktriangle)}$ & $0.932 \rightarrow 0.725\,{\scriptstyle (\blacktriangle)}$ & 0.899 & $0.256 \rightarrow 0.668\,{\scriptstyle (\blacktriangle)}$ & $0.965 \rightarrow 0.792\,{\scriptstyle (\blacktriangle)}$ \\
AT & 0.928 & $0.676 \rightarrow 0.842\,{\scriptstyle (\bigstar)}$ & $0.758 \rightarrow 0.758\,{\scriptstyle (\text{\ppabar})}$ & 0.818 & $0.415 \rightarrow 0.569\,{\scriptstyle (\bigstar)}$ & $0.822 \rightarrow 0.822\,{\scriptstyle (\text{\ppabar})}$ \\
AT* & 0.891 & $0.430 \rightarrow 0.632\,{\scriptstyle (\bigstar)}$ & $0.849 \rightarrow 0.829\,{\scriptstyle (\bigstar)}$ & 0.748 & $0.266 \rightarrow 0.364\,{\scriptstyle (\bigstar)}$ & $0.887 \rightarrow 0.887\,{\scriptstyle (\text{\ppabar})}$ \\
\bottomrule
\end{tabular}
\end{adjustbox}
\par\smallskip
{\footnotesize PPA (baseline): $\text{\ppabar}$\qquad
PPA N-LR: $\bigstar$\qquad PPA C-LR: $\blacktriangle$}
\end{table}

\begin{table}[t]
\centering
\small
\setlength{\tabcolsep}{3pt}
\renewcommand{\arraystretch}{1.05}
\caption{ResNet-152 with standard and robust Inception evaluation. Each pair shows
AttAcc@$k$ $\rightarrow$ Robust-AttAcc@$k$ for $k\in\{1,5\}$ from the same
vanilla PPA run. Bold values mark changes $>$$0.1$.}
\label{tab:resnet152-ppa-standard-robust}
\begin{adjustbox}{max width=\linewidth}
\begin{tabular}{@{}lccc@{\hspace{12pt}}ccc@{}}
\toprule
& \multicolumn{3}{c}{\textbf{FaceScrub}} & \multicolumn{3}{c}{\textbf{CelebA}} \\
\cmidrule(lr){2-4}\cmidrule(lr){5-7}
\textbf{Model} & \textbf{TestAcc} $\uparrow$ &
\textbf{AttAcc@1} $\downarrow$ &
\textbf{AttAcc@5} $\downarrow$ &
\textbf{TestAcc} $\uparrow$ &
\textbf{AttAcc@1} $\downarrow$ &
\textbf{AttAcc@5} $\downarrow$ \\
\midrule
NoDef & 0.943 & \textbf{0.823} $\rightarrow$ \textbf{0.463} & \textbf{0.961} $\rightarrow$ \textbf{0.761} & 0.845 & 0.459 $\rightarrow$ 0.518 & 0.750 $\rightarrow$ 0.786 \\
\midrule
MID & 0.950 & \textbf{0.391} $\rightarrow$ \textbf{0.219} & \textbf{0.699} $\rightarrow$ \textbf{0.489} & 0.802 & 0.474 $\rightarrow$ 0.441 & 0.725 $\rightarrow$ 0.688 \\
BiDO & 0.920 & \textbf{0.754} $\rightarrow$ \textbf{0.494} & \textbf{0.937} $\rightarrow$ \textbf{0.781} & 0.747 & 0.362 $\rightarrow$ 0.387 & 0.629 $\rightarrow$ 0.653 \\
TL\mbox{-}DMI & 0.891 & 0.151 $\rightarrow$ 0.131 & 0.384 $\rightarrow$ 0.318 & 0.814 & 0.281 $\rightarrow$ 0.311 & 0.550 $\rightarrow$ 0.591 \\
RoLSS & 0.886 & 0.474 $\rightarrow$ 0.470 & 0.760 $\rightarrow$ 0.777 & 0.537 & 0.046 $\rightarrow$ 0.066 & 0.145 $\rightarrow$ 0.182 \\
RoLSS\mbox{-}SSF & 0.869 & 0.343 $\rightarrow$ 0.350 & 0.635 $\rightarrow$ 0.653 & 0.428 & 0.039 $\rightarrow$ 0.030 & 0.118 $\rightarrow$ 0.097 \\
MIDRE & 0.941 & \textbf{0.456} $\rightarrow$ \textbf{0.343} & \textbf{0.761} $\rightarrow$ \textbf{0.624} & 0.756 & 0.167 $\rightarrow$ 0.223 & 0.377 $\rightarrow$ 0.461 \\
Trap\mbox{-}MID & 0.926 & \textbf{0.401} $\rightarrow$ \textbf{0.232} & \textbf{0.676} $\rightarrow$ \textbf{0.481} & 0.806 & 0.092 $\rightarrow$ 0.097 & 0.193 $\rightarrow$ 0.195 \\
NegLS & 0.911 & 0.105 $\rightarrow$ 0.069 & 0.277 $\rightarrow$ 0.200 & 0.805 & 0.340 $\rightarrow$ 0.333 & 0.623 $\rightarrow$ 0.639 \\
\midrule
AT\mbox{-}AT & 0.941 & 0.058 $\rightarrow$ 0.042 & 0.190 $\rightarrow$ 0.145 & 0.819 & 0.049 $\rightarrow$ 0.059 & 0.138 $\rightarrow$ 0.151 \\
\midrule
MixUp & 0.971 & \textbf{0.488} $\rightarrow$ \textbf{0.331} & \textbf{0.727} $\rightarrow$ \textbf{0.588} & 0.899 & 0.256 $\rightarrow$ 0.294 & 0.483 $\rightarrow$ 0.511 \\
AT & 0.928 & \textbf{0.676} $\rightarrow$ \textbf{0.934} & 0.896 $\rightarrow$ 0.995 & 0.818 & \textbf{0.415} $\rightarrow$ \textbf{0.713} & \textbf{0.693} $\rightarrow$ \textbf{0.912} \\
AT* & 0.891 & \textbf{0.430} $\rightarrow$ \textbf{0.901} & \textbf{0.715} $\rightarrow$ \textbf{0.982} & 0.748 & \textbf{0.266} $\rightarrow$ \textbf{0.734} & \textbf{0.527} $\rightarrow$ \textbf{0.929} \\
\bottomrule
\end{tabular}
\end{adjustbox}
\end{table}

\begin{figure}[h]
    \centering
    \includegraphics[width=1\linewidth]{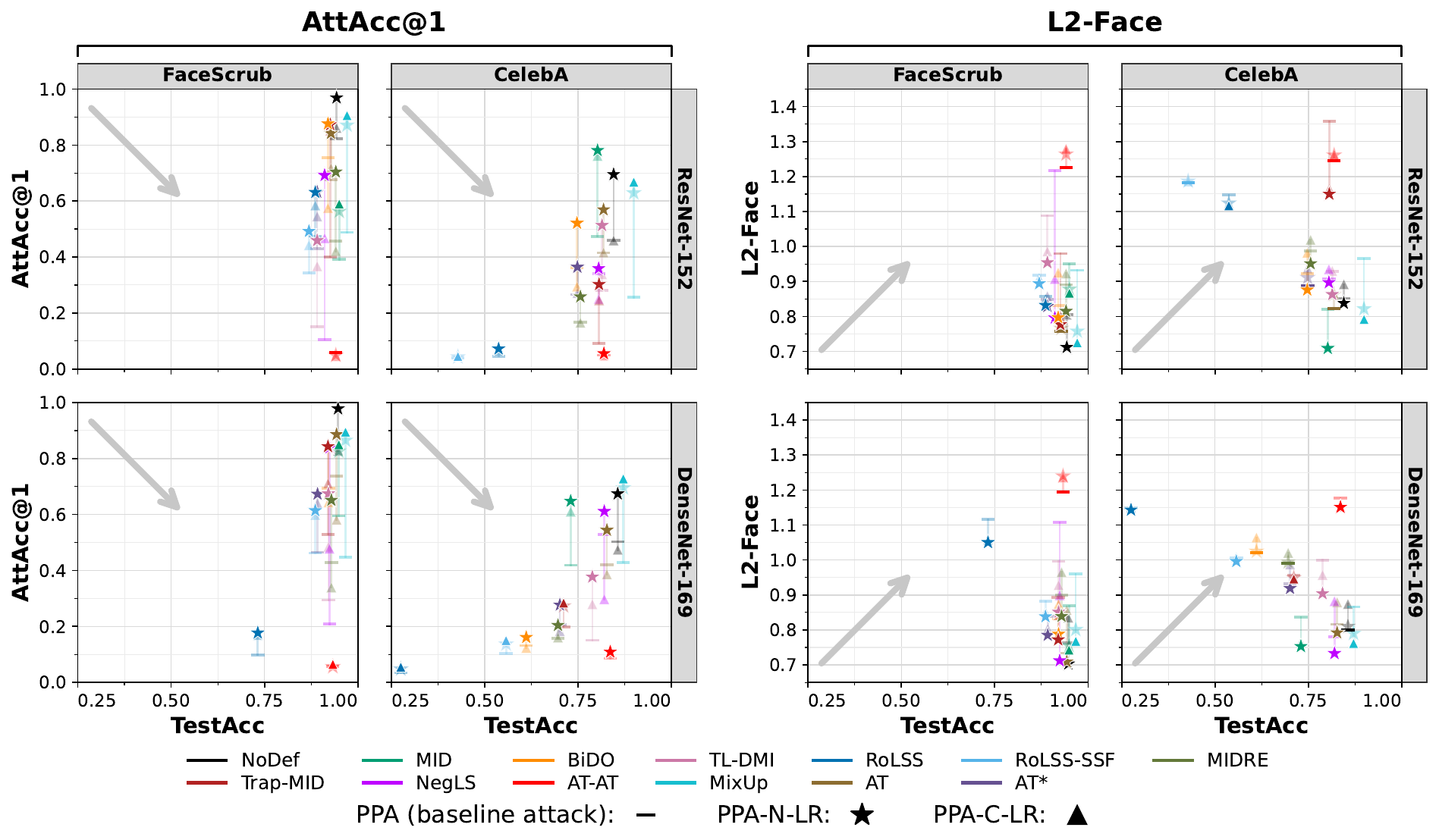}
    \caption{Leakage (AttAcc@1 \& L2-Face) vs. test accuracy. Shapes denote attacks. Vertical lines measure vanilla PPA vs. the adaptive attacks, and opaque markers highlight the worst-case leakage. }
    \label{fig:results-adaptive-attacks}
\end{figure}

\textbf{Results.} Fig. \ref{fig:results-adaptive-attacks} plots the performance of each defense under both unmodified SOTA PPA attacks and PPA attacks adapted with our techniques. Table \ref{tab:resnet152-ppa-worst-case} summarizes each defense's worst-case increase in leakage under our techniques as compared to vanilla PPA. On the defenses widely regarded as SOTA in terms of performance, AttAcc@1 leakage increases from, e.g., $0.105$ to $0.692$ (NegLS on FaceScrub under RN-152), or from $0.401$ to $0.873$ (Trap-MID on FaceScrub under RN-152). In terms of generic techniques, MixUp increases from, e.g., $0.256$ to $0.668$ (on CelebA under RN-152). For RN-152, AttAcc@1 exceeds standard PPA on 25 of 26 model--dataset pairs. App. \ref{deferred-model-inversion-results} shows full results, plus results applying our techniques to IF-GMI attacks. \textbf{Code is on \href{https://github.com/ShailenSmith/Adaptive-MIA}{Github}.}

We note that these increases in leakage are substantial. For comparison, on FaceScrub under our techniques, SOTA defense Trap-MID leaks more in terms of AttAcc@1  than an undefended model leaks under vanilla PPA. Our results therefore indicate that a substantial share of reported defense performance was overfit against one fixed PPA configuration. Interestingly, the lone `defense' little-affected by our techniques is AT-AT, which is not a practical defense, but a tool to show privacy gain obtainable by deliberately sacrificing the model's adversarial robustness \citep{torp2026reducing}.

\textbf{Robust evaluation results.} Table \ref{tab:resnet152-ppa-standard-robust} and App. \ref{deferred-model-inversion-results} recompute the 2 metrics that depend on an external Inception-v3 model (AttAcc@1, AttAcc@5) using a robust AT-trained version of this model, holding the attacker reconstructions of each class fixed. For the higher-$\epsilon$ FaceScrub AT model, standard PPA obtains AttAcc@1 of $0.430$ under the conventional evaluator and $0.901$ under the robust evaluator.
\newpage
C-LR rises from $0.545$ to $0.960$ under the same evaluator substitution, and N-LR from $0.632$ to $0.942$. These results indicate that standard AttAcc@1 used as the main privacy metric in the high-resolution literature significantly understates leakage precisely where the target model relies most on robust, visually meaningful features.
\section{Generalizing the Privacy--Robustness Tradeoff}
\label{sec:results-regressions}
\begin{figure}[t]
    \centering
    \includegraphics[width=0.75\linewidth]{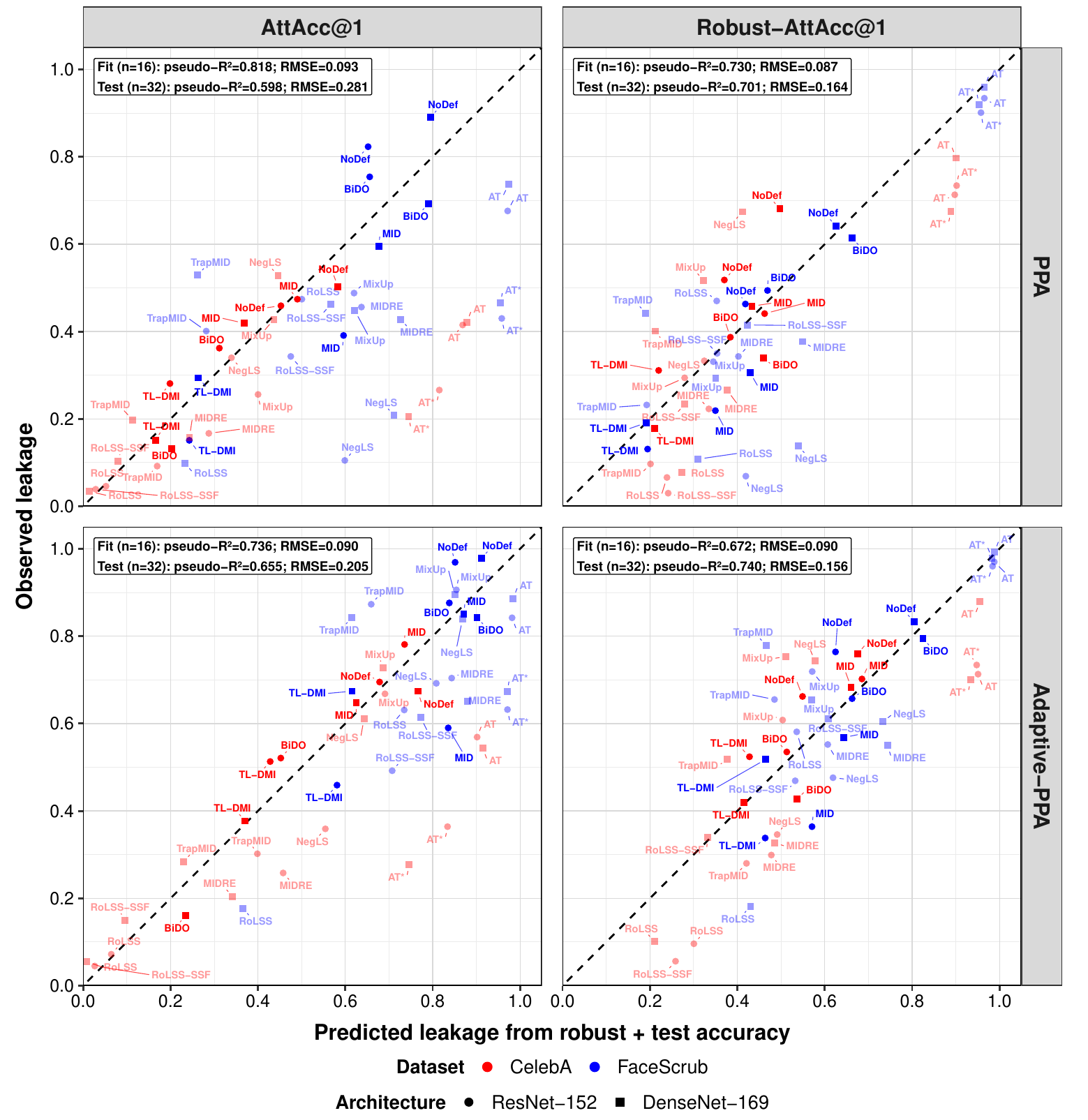}
\caption{The privacy--robustness relationship generalizes substantially better under adaptive attacks (bottom row) and a Robust Inception-v3 evaluator (right column).
Opaque points were used for fitting (the same set of defenses that were used to fit in the \citet{torp2026reducing} regression).}
    \label{fig:regression-combined}
    \vspace{-2em}
\end{figure}

We now show that the privacy--robustness tradeoff \citep{torp2026reducing} applies generally across defenses and generalization training techniques in the field rather than to just a small set of 3 generic defenses, and importantly, the tradeoff does not exclude high-performance anti-gradient defenses as theorized in \citet{torp2026reducing}, or generalization training techniques widely used by practitioners. Fig. \ref{fig:regression-combined} plots the fit of the same leakage-vs-robustness  regression as \citet{torp2026reducing}, run on the same in-model defenses (opaque points: NoDef, BiDO, MID, TL-DMI). Unlike \citet{torp2026reducing}, we also plot high-performance and anti-gradient defenses (NegLS, Trap-MID, RoLSS, RoLSS-SSF, MIDRE) and generalization techniques (MixUp, AT, and AT*) as an out-of-sample test-set (semi-transparent points). 
We exclude AT-AT \citep{torp2026reducing} here,  as it is a causal instrument and not a practical defense, and it deliberately optimizes the non-robustness predictor in the regressions (Fig.~\ref{fig:adaptive_all_defenses_combined} in the Appendix refits on all 52 models including AT-AT (pseudo-$R^2$ 0.697 and 0.773)). 
Top-row regressions show the regression model using vanilla PPA as in  \citet{torp2026reducing}, and the bottom-row recomputes the regressions using the worst-case over the adaptive attacks and PPA. The left column uses vanilla AttAcc@1, and the right column recomputes with Robust AttAcc@1 under the AT-trained Inception v3 variant. Thus, the bottom-right regression panel combines both changes.

Each change improves privacy--robustness leakage prediction on  held-out defenses. Under vanilla PPA and vanilla evaluation (top left), the regression obtains test pseudo-$R^2$$=$$0.598$, RMSE $=$$0.281$. Adaptive attacks raise pseudo-$R^2$ to $0.655$ and lower RMSE to $0.205$. Robust evaluation raises pseudo-$R^2$ to $0.701$ and lowers RMSE to $0.164$. Combining adaptive attacks (bottom) and robust evaluation (right) produces the strongest held-out fit: $R^2$=$0.740$, RMSE $=$$0.156$ (bottom-right).

The original privacy--robustness tradeoff was demonstrated only on 3  generic defenses and theorized not to apply to recent gradient-blocking ones \citep{torp2026reducing}. Here under adaptive attack and robust evaluation, it generalizes to the tested recent defenses and to both generic training techniques, without adding any of them to the fitting set. Thus, the privacy--robustness tradeoff generalizes when measured by attacks and evaluators to which defenses are not overfit. We also emphasize that these substantial improvements to the fit were made with a small set of simple attack adaptations and a single more robust evaluation variant. We hypothesize that as adaptive attacks improve to better approximate the optimal reconstructions, the privacy--robustness fit will improve accordingly.

\section{Discussion}
\label{sec:discussion}
The improved out-of-sample fit suggests a generalizable empirical relationship between adversarial robustness and privacy leakage that applies more broadly than previously theorized. This poses a key question: why should robustness predict privacy?

\textbf{From Robustness to Gradient Alignment.}
Recent work on input gradient geometry suggests a possible explanation. \citet{srinivas2023models} argue that input gradients $\nabla_x{}T(x)$ are \textit{perceptually aligned} when they lie on the signal manifold containing the task-relevant components of the data distribution. They further show that for Bayes-optimal classifiers, such on-manifold alignment is equivalent to off-manifold robustness, and show empirically that off-manifold robustness is prevalent across several forms of robust training. %
In this way, prior work connects robustness and perceptual alignment in both directions: robust models tend to exhibit perceptually aligned gradients \citep{srinivas2023models,kaur2019perceptually}, while directly promoting perceptual alignment can itself increase robustness \citep{ganz2023perceptually}. Separately, \citet{peng2026generative} show that generative MIAs implicitly project inversion-time loss gradients onto the tangent space of the generator manifold, filtering out components that the generator cannot follow. They empirically show that models become more vulnerable to inversion when their loss gradients are better aligned with that manifold.

These are different notions of alignment. The signal manifold is defined by the task and data distribution; the generator manifold is defined by the attacker's auxiliary generator. However, when the generator captures the relevant semantic structure of the private distribution, the two results suggest a natural connection. Robust models concentrate more of their predictive gradients on semantically meaningful directions. If those directions are also represented by the generator, then it follows that more of the target model's gradient is available to a generator-constrained attack. In contrast, non-robust features remain generally predictive, but are visually imperceptible and need not provide equally useful directions for reconstructing recognizable identity. This provides one geometric explanation for why adversarial robustness predicts reconstruction leakage.

\citet{peng2026generative} quantify their mechanism using an inversion-time alignment score; Fig. \ref{fig:regression-alignment} regresses this alignment score with AttAcc@1 across the same 48 target models shown in Fig. \ref{fig:regression-combined}. We find the expected positive relationship with leakage, but considerably weaker predictive performance than adversarial robustness: the alignment regression obtains pseudo-$R^2=0.244$ on the fitting set and $0.085$ across the test set. This reflects an important distinction: Alignment is measured along a particular attack trajectory and therefore depends on the generator, loss, current reconstruction, and attack optimization. In contrast, robustness is measured independently of the inversion attack. In our experiments, it provides a stronger attack-agnostic proxy for reconstruction vulnerability, whereas alignment helps explain the mechanism by which  inversion reconstructs  data.

\begin{figure}[H]
\vspace{-0.5em}
  \centering
  \includegraphics[width=\linewidth]{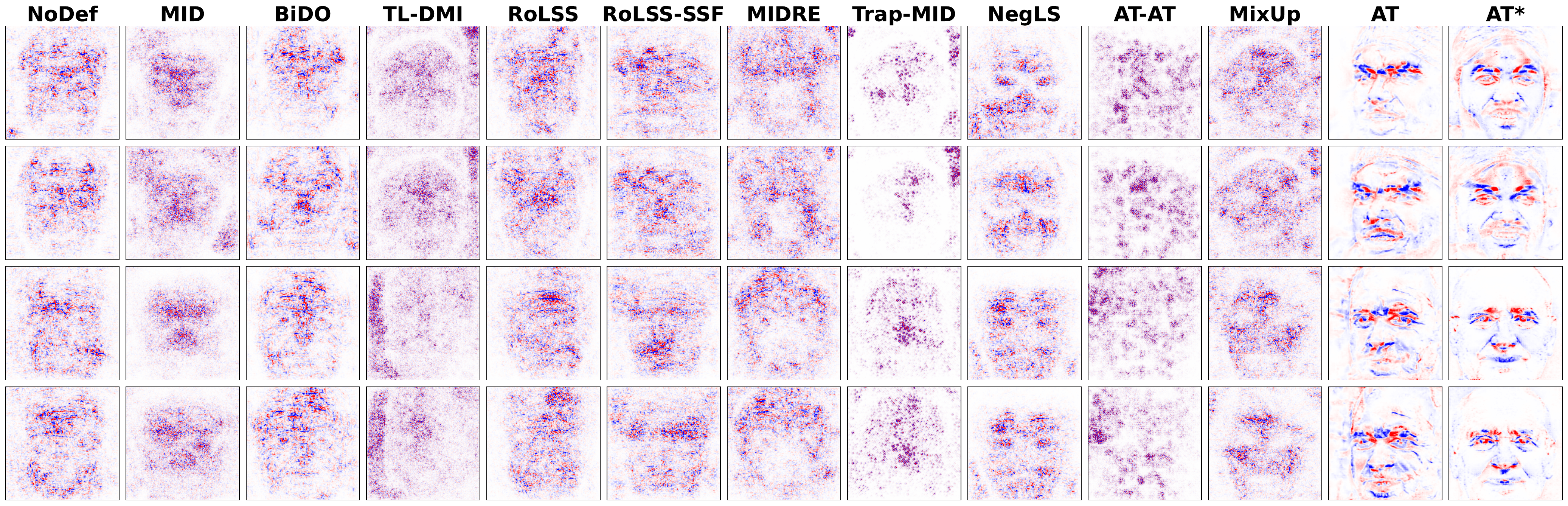}
  \caption{Heatmaps of inversion-time loss gradients $\nabla_x\mathcal{L}_\text{cls}(x)$, where $x=G(z)$ is a GAN-generated reconstruction at various PPA iterations on ResNet-152. Gradient magnitudes are \textit{normalized}, i.e., incomparable across target models. Rows 1 \& 2 correspond to FaceScrub class 12 at attack iterations 9 and 49, respectively. Rows 3 \& 4 correspond to the same iterations of class 16.}
  \label{fig:gradient-heatmaps}
  \vspace{-1em}
\end{figure}

\textbf{Semantic Structure in Inversion Gradients.}
Interestingly, the same relationship is visible in the inversion gradients themselves. Fig.~\ref{fig:gradient-heatmaps} plots normalized vanilla-PPA gradients across the target models in our experiments, as done in \citet{peng2026generative}, holding the attack, generator, architecture, and dataset fixed. The plots therefore  visualize the gradient structure exposed by each target model under the same inversion attack. Across the target models, greater adversarial robustness generally corresponds to more recognizable structure in the inversion gradients. The adversarially trained models provide the clearest examples: across both identities and attack iterations, their gradients trace the eyes, nose, mouth, and contours of the face. This illustration is consistent with prior work showing that robust classifiers produce perceptually aligned gradients~\citep[e.g.,][]{tsipras2018robustness,kaur2019perceptually,ganz2023perceptually, srinivas2023models}.

More surprisingly, low leakage under vanilla PPA does not imply the absence of semantically structured gradients. NegLS provides the clearest example: vanilla PPA obtains AttAcc@1 of $0.105$, yet its gradients visibly trace the eyes, nose, and mouth across all four panels; our adaptive attack raises leakage to $0.692$. TL-DMI and RoLSS-SSF exhibit weaker but still recognizable facial structure, and their leakage likewise increases under adaptive PPA. Trap-MID illustrates the complementary case. Its vanilla gradients are sparse and substantially less facial, yet adaptive PPA raises AttAcc@1 from $0.401$ to $0.873$. The heatmaps therefore cannot by themselves measure reconstruction vulnerability. They describe the gradients encountered by one attack trajectory. A fixed attack may fail despite receiving semantically useful gradients, or may follow a trajectory that never exposes the gradients available under another loss. In either case, low measured leakage under one attack does not establish that the recoverable information has been removed.

\section{Replication \& Code}
The experiments in this paper were designed from the outset to be replicated. We provide full replication code for the attacks, evaluations, regressions, and experiments reported in this paper: \  \href{https://github.com/ShailenSmith/Adaptive-MIA}{\textbf{GitHub}}. \   Wherever possible, we use the original authors' implementations and parameters for MIAs and defenses; see App. \ref{app:sec-experimental-details} for full training and attack experimental details.

App.~\ref{app:regression-details}  provides regression specifications and coefficients, and App. \ref{deferred-model-inversion-results} reports full experimental results across attacks, defenses, datasets, architectures, and evaluation metrics. Our code is also organized so that the adaptive attacks and evaluation methods introduced here can be applied directly to additional MIA defenses.

\section{Conclusion}
\label{sec:conclusion}
In this paper, we show that SOTA privacy defenses, generic training techniques intended to improve generalization, and undefended models all expose their training data more, and in some cases several fold more, than reported in standard MIA evaluations. Simple changes to the attack loss and learning rate recover substantially more leakage, while scoring the same reconstructions under an alternative evaluator feature basis can materially change measured leakage. This is a familiar failure mode from adversarial robustness: defenses can appear effective because they overfit to the standard evaluation configuration, rather than because they mitigate the underlying vulnerability. These underestimated leakage rates also concealed a more general relationship; once we adapt the attack and vary the evaluator, the privacy--robustness relationship generalizes substantially better to held-out defenses and standard training regimes that were excluded from the original fit. Adversarial robustness therefore provides an attack-agnostic proxy for reconstruction vulnerability in this setting. It remains an open question whether any practical defense can beat this privacy--robustness curve by reducing training-data exposure without paying the corresponding cost in adversarial robustness.

\section{Acknowledgments}
Breuer Lab gratefully acknowledges the support of the  \textls[120]{OpenAI}  \textsc{Cybersecurity Grant.}

\clearpage

\bibliography{references}
\bibliographystyle{plainnat}

\appendix

\clearpage

\section{Deferred Details of Experiments}
\label{app:sec-experimental-details}

\subsection{Datasets}

\begin{itemize}
\item \textbf{FaceScrub \citep{ng_data-driven_2014}.}
FaceScrub contains 530 celebrity identities and is widely used in model inversion research \citep{struppek_be_2024, koh_vulnerability_2024, ho_model_2024, qiu_closer_2024, struppek_plug_2022}. Following prior work, we use a commonly used Kaggle-hosted version of the dataset\footnote{\url{https://www.kaggle.com/datasets/rajnishe/facescrub-full}}.

\item \textbf{CelebA \citep{liu_deep_2015}.} 
CelebA contains 202,599 images spanning 10,177 celebrity identities. Following prior model inversion work \citep{struppek_be_2024, liu2024trapmid, qiu_closer_2024, struppek_plug_2022}, we retain the 1,000 identities with the most images, yielding 30,038 images. We additionally process the images using the CelebA HD Cropper\footnote{\url{https://github.com/LynnHo/HD-CelebA-Cropper}}.
\end{itemize}

\subsection{Model Architectures}
We evaluate two architectures commonly used in recent high-resolution model inversion work:

\begin{itemize}
    \item \textbf{ResNet-152 \citep{he_deep_2015}.}
    ResNet-152 is widely used in prior MIA studies \citep{struppek_plug_2022, qiu_closer_2024, koh_vulnerability_2024, struppek_be_2024} and serves as our primary architecture.
    \item \textbf{DenseNet-169 \citep{huang_densely_2017}.}
    We additionally evaluate DenseNet-169, following prior MIA work \citep{struppek_plug_2022, qiu_closer_2024, koh_vulnerability_2024, liu2024trapmid}.
\end{itemize}

\subsection{Target Models}
\label{app:defenses-deferred-experimental-setting}

To reduce the computational and environmental impact of this paper, we reached out to authors of \citet{torp2026reducing} for access to their target model weights. We note here which ones we were able to use and which we have trained from scratch in this paper: 
\begin{itemize}
    \item \textbf{No Defense (NoDef).} We use a vanilla ResNet-152 and DenseNet-169 trained without any model inversion defense as the baseline.
    \item \textbf{Mutual Information Defense (MID) \citep{wang_improving_2020}.} The MID defense utilizes a regularizer that has the goal of reducing training-data-to-model information dependency. It does so by penalizing an approximation of the mutual information of training data and intermediate feature representations. We get the trained weights from \citet{torp2026reducing}.  
    \item \textbf{Bilateral Dependency Optimization (BiDO) \citep{peng_bilateral_2022}.} BiDO uses two regularizers: one reduces a kernel-based dependency measure between inputs and intermediate representations to limit information flow from the training data, while the other increases dependency between intermediate representations and outputs to preserve predictive performance. We get the trained weights from \citet{torp2026reducing}.
    \item \textbf{Transfer Learning Defense against Model Inversion (TL-DMI) \citep{ho_model_2024}.} Most defenses train the target model by fully fine-tuning a publicly pre-trained model on the private train set. In contrast, TL-DMI fine-tunes only the deeper layers, with the goal of reducing the number of parameters that encode sensitive information from the private dataset. We train it for ResNet-152 FaceScrub, freezing the first $6$ layers. For all other model architecture/defense combinations we get the weights from \citet{torp2026reducing}.
    \item \textbf{Negative Label Smoothing (NegLS) \citep{struppek_be_2024}.} NegLS applies negative label smoothing to make the target model systematically overconfident in its predictions. Rather than removing training-data information from the model, this deliberate overconfidence disrupts the confidence-score landscape used by PPA and related gradient-based GAN attacks, such as IF-GMI, making the gradients used for reconstruction less effective. 
    \item \textbf{Removal of Last Stage Skip-Connection (RoLSS) \citep{koh_vulnerability_2024}.}  RoLSS removes the skip connection in the final stage of the target model to disrupt gradient flow during the optimization of PPA and related attacks. We get the trained weights from \citet{torp2026reducing}.
    \item \textbf{Skip-Connection Scaling Factor (RoLSS-SSF) \citep{koh_vulnerability_2024}.} RoLSS-SSF reduces, rather than removes, the contribution of the final-stage skip connection, aiming to impede gradient flow for inversion attacks while retaining more of the original model architecture and predictive performance. We get the trained weights from \citet{torp2026reducing}.
    \item \textbf{Trapdoor-based Defense (Trap-MID)
    \citep{liu2024trapmid}.} Trap-MID embeds a label-specific trapdoor into the target model, causing inputs containing the corresponding trigger to be classified as that label. This trapdoor acts as a shortcut for model inversion attacks, steering reconstructions toward the trigger pattern rather than private training data. We use the models from \citet{torp2026reducing}, which reports two hyperparameter variants for each dataset--architecture pair, and retain only the more private variant for each dataset: $(\alpha=0.007,\beta=0.5)$ for FaceScrub and $(\alpha=0.02,\beta=0.2)$ for CelebA.
    \item \textbf{Model Inversion Defense via Random Erasing (MIDRE) \citep{tran2025random}.} For the MIDRE defense, the authors argue for a target model trained on randomly partially erased images, limiting its exposure to complete objects. This aims to create a mismatch between private-data features and MI reconstructions. We train the MIDRE defense with $\alpha=0.4$, which denotes the range of random erasing. The implementation uses \texttt{Torchvision RandomErasing} in the same manner as the original authors. 
\end{itemize}
Although originally presented by \citet{torp2026reducing} as a causal mechanism to demonstrate a privacy-robustness tradeoff, and not as a model inversion defense, we also evaluate AT-AT in our model inversion attack experiments. 
\begin{itemize}
    \item \textbf{Anti-Adversarial Training (AT-AT) \citep{torp2026reducing}.} The AT-AT training regime has the opposite goal of ordinary adversarial training, that is to be as non-robust as possible under even very small magnitude perturbations, while still retaining good generalization on clean data. This is achieved by balancing a loss of vanilla cross-entropy on the label and cross-entropy of a targeted PGD attack with respect to a random target class. 
\end{itemize}

We additionally consider these commonly used training setups that we find to greatly affect the standard model inversion attack setups:
\begin{itemize}
    \item \textbf{Adversarial Training (AT) \citep{madry_towards_2017}.}
    AT trains the target model on adversarially perturbed inputs, encouraging robustness within a bounded perturbation region and reducing reliance on non-robust features. AT and AT$^\ast$ use $\epsilon=0.01$ and $0.031$ on FaceScrub, and $\epsilon=0.0025$ and $0.01$ on CelebA, respectively; AT$^\ast$ always denotes the higher $\epsilon$. All AT models are trained with 10-step PGD under an $\ell_\infty$ perturbation constraint.
    \item \textbf{MixUp \citep{zhang2018mixup}.} Mixup trains the target model on convex combinations of pairs of training examples and their labels, encouraging smoother behavior between samples and reducing memorization. It is a widely used data augmentation technique for its generalization improvements. We use Mixup with $\alpha=0.4$.
\end{itemize}

\subsection{Target Model Training}
Unless otherwise specified, we follow the target-model training setup of \citet{struppek_be_2024, struppek_plug_2022}. Models are initialized from ImageNet-pretrained Torchvision weights and trained for 100 epochs using Adam with $\beta=(0.9,0.999)$, batch size 128, and an initial learning rate of $0.001$. We use a MultiStepLR scheduler with $\gamma=0.1$ and retain the checkpoint with the lowest validation loss. Final prediction accuracy is reported on the test split.

\subsection{Evaluation Model Training}
\begin{itemize}
    \item \textbf{Vanilla evaluation model.}
Following \citet{struppek_plug_2022}, we use independently trained Inception-V3 models for evaluation. The CelebA Inception model is hot-started from the FaceScrub Inception model to improve accuracy, as is standard \citep{struppek_plug_2022}. The FaceScrub and CelebA vanilla Inception models achieve test accuracies of 96.9\% and 91.2\%, respectively.
    \item \textbf{Robust evaluation model.}
We additionally train an adversarially robust evaluation model using 10-step PGD adversarial training with an $\ell_\infty$ perturbation budget of $\epsilon=0.031$ for both FaceScrub and CelebA. The FaceScrub and CelebA robust Inception models achieve test accuracies of 93.3\% and 83.4\%, respectively.
\end{itemize}

\subsection{Model Inversion Attacks}\label{app:attacks}
\begin{itemize}
    \item \textbf{Plug-and-Play Attack (PPA) \citep{struppek_plug_2022}.}
    PPA performs inversion in three stages: latent sampling, latent optimization, and final sample selection. It uses a StyleGAN-2 \citep{karras_analyzing_2020} generator pre-trained on FFHQ \citep{karras2019stylebasedgeneratorarchitecturegenerative}, which has no identity overlap with FaceScrub or CelebA. We evaluate PPA on 100 target classes with a batch size of 40. Latent codes are optimized for 50 steps, and otherwise we follow the standard hyperparameter setup of \citet{struppek_plug_2022, struppek_be_2024}.

    \item \textbf{Intermediate Features enhanced Generative Model Inversion (IF-GMI) \citep{qiu_closer_2024}.}
    IF-GMI builds on GAN-based inversion by additionally optimizing intermediate StyleGAN-2 features, increasing the flexibility of the reconstructed images. We use the same FFHQ-pretrained StyleGAN-2 prior and evaluate the attack on 100 target classes. We otherwise follow the hyperparameter settings of the original implementation.
\end{itemize}

\subsubsection{Adversarial Attacks}
\begin{itemize}
    \item \textbf{AutoAttack \citep{croce_reliable_2020}}. Following \citep{torp2026reducing}, we measure adversarial robustness of target models using the AutoAttack ensemble of 4 attacks (APGD-CE, APGD-T, FAB-T, and Square) meant to provide a thorough, diverse signal of robustness with minimal hyperparameter tuning \citep{croce_reliable_2020, croce_minimally_2020, andriushchenko_square_2020}. We use an $l_\infty$ norm and measure across various $\epsilon$ levels $\{0.031, 0.0025, 0.0005, \text{1e-5}\}$. We exclude $\epsilon=0.031$ from the regression as it is less informative in these experimental settings.
\end{itemize}

\subsection{Alignment Score}
In order to quantify the notion of inversion-time loss gradient alignment, \citet{peng2026generative} propose an \textit{alignment score} measure, defined as the cosine of the angle $\phi$ between a loss gradient $\nabla_x{}\mathcal{L}_\text{cls}$ and its projection onto the tangent space $T_x\mathcal{M}_\text{aux}$ of the GAN manifold at point $x$:
$
   \cos(\phi) = \left\| \mathbf{P}_{\mathbf{x}} \nabla_{\mathbf{x}} \mathcal{L}_{\text{cls}} \right\| \,/\, \left\| \nabla_{\mathbf{x}} \mathcal{L}_{\text{cls}} \right\|,
$
where $P_x$ is an orthogonal projection matrix onto the tangent space $T_x\mathcal{M}_\text{aux}$. Higher alignment scores are hypothesized to correspond to increased MIA vulnerability.

One alignment score can be calculated per sample per attack iteration. Since alignment scores are very computationally costly to calculate, we must subsample. We compute on every tenth iteration (6 iterations total), a subset of 15 classes, and just 1 sample per 40-size batch. With 5 batches per class, this leads to 450 alignment scores per attack; we report the mean of these scores in the tables and regressions.

\subsection{Compute Setup}
We run most target model training and attacks on NVIDIA Blackwell Pro 6000 Max-Q, NVIDIA A100, and NVIDIA H100 GPUs. Default target model training and attack hyperparameters are set up to fit on a single 80 GB GPU.

\section{Deferred Regression Details}
\label{app:regression-details}

We use the same Beta-regression specification as \citet{torp2026reducing}, using the mean--precision parameterization of \citet{ferrari2004beta}. For target-model condition $j$, let $Y_j\in(0,1)$ denote measured reconstruction leakage, $\mathrm{TestAcc}_j$ denote clean test accuracy, and $\mathrm{AutoAcc}_{\epsilon,j}$ denote AutoAttack robust accuracy at perturbation magnitude $\epsilon$. We estimate
\begin{align}
Y_j &\sim \mathrm{Beta}(\mu_j\phi,(1-\mu_j)\phi), \nonumber\\
\mathrm{logit}(\mu_j)
&=
\beta_0
+\beta_1\mathrm{TestAcc}_j
+\beta_2\mathrm{AutoAcc}_{0.0025,j}
+\beta_3\mathrm{AutoAcc}_{0.0005,j}
+\beta_4\mathrm{AutoAcc}_{10^{-5},j}.
\label{eq:privacy_robustness_regression}
\end{align}
As in \citet{torp2026reducing}, clean test accuracy controls for the known relationship between model accuracy and MIA leakage, while the three AutoAttack terms measure robustness to small adversarial perturbations. We omit $\epsilon=0.031$, at which robust accuracy is near zero for nearly all non-adversarially-trained target models.

For the main regressions in Fig.~\ref{fig:regression-combined}, we fit Eq.~\ref{eq:privacy_robustness_regression} only on NoDef, MID, BiDO, and TL-DMI across FaceScrub and CelebA and ResNet-152 and DenseNet-169 ($n=16$). NegLS, Trap-MID, MIDRE, RoLSS, RoLSS-SSF, MixUp, AT, and AT* are held out entirely ($n=32$). AT-AT is excluded because it is a causal instrument deliberately constructed to manipulate the non-robust feature basis captured by the robustness predictors (we also re-run the regression with AT-AT included, as described below).

We estimate four regressions, changing only the leakage response $Y_j$: AttAcc@1 under vanilla PPA; Robust-AttAcc@1 on the same vanilla-PPA reconstructions; the maximum AttAcc@1 across PPA, N-LR, and C-LR; and the maximum Robust-AttAcc@1 across those attacks. The two maxima are computed independently. Predictors, fitting observations, and held-out observations are otherwise identical across all four regressions. Table~\ref{tab:leakage-regression} reports the fitted coefficients and fit statistics.

Three additional regression figures use related specifications. Fig.~\ref{fig:regression-outlier} applies the published coefficients of \citet{torp2026reducing}, fit on their $n=18$ observations, to the $n=32$ held-out observations in this paper without refitting. Fig.~\ref{fig:regression-alignment} replaces the robustness predictors with the vanilla-PPA alignment score as the predictor of AttAcc@1, while retaining the same $n=16$ fitting and $n=32$ held-out observations. Finally, Fig.~\ref{fig:adaptive_all_defenses_combined} refits Eq.~\ref{eq:privacy_robustness_regression} on all $52$ target-model conditions, including AT-AT, with no held-out set.

\textbf{Table \ref{tab:leakage-regression}} reports $\beta$ coefficients for the four main body regressions (Fig. \ref{fig:regression-combined}).

\begin{table}[t]
\centering
\begin{compacttable}
\caption{Beta regression (logit link) coefficients predicting leakage from test and
AutoAttack robust accuracy, with pseudo-$R^2$ and RMSE on the fitted and held-out models.}
\label{tab:leakage-regression}
\setlength{\tabcolsep}{3pt}
\begin{adjustbox}{max width=\linewidth}
\begin{tabular}{@{}l*{4}{c}@{}}
\toprule
 & \multicolumn{2}{c}{\textbf{PPA}} & \multicolumn{2}{c}{\textbf{Adaptive PPA}} \\
\cmidrule(lr){2-3}\cmidrule(lr){4-5}
\textbf{Term} & \textbf{AttAcc@1} & \textbf{Robust-AttAcc@1} & \textbf{AttAcc@1} & \textbf{Robust-AttAcc@1} \\
\midrule
Intercept & $-5.836$ & $-0.970$ & $-7.303$ & $-1.917$ \\
TestAcc & $6.422$ & $-0.208$ & $11.005$ & $3.142$ \\
AutoAcc.0025 & $3.434$ & $4.271$ & $2.736$ & $4.662$ \\
AutoAcc.0005 & $1.608$ & $1.060$ & $1.132$ & $0.429$ \\
AutoAcc1e-5 & $-1.222$ & $-0.316$ & $-2.602$ & $-1.229$ \\
\midrule
Fit ($n=16$) pseudo-$R^2$ & 0.818 & 0.730 & 0.736 & 0.672 \\
Fit ($n=16$) RMSE & 0.093 & 0.087 & 0.090 & 0.090 \\
Test ($n=32$) pseudo-$R^2$ & 0.598 & 0.701 & 0.655 & 0.740 \\
Test ($n=32$) RMSE & 0.281 & 0.164 & 0.205 & 0.156 \\
\bottomrule
\end{tabular}
\end{adjustbox}
\end{compacttable}
\end{table}

\section{Additional Related Work}

\subsection{Adaptive Attacks under MIA}
\label{app:related-work-adaptive-mia}
While white-box model inversion has lacked a consistent, rigorous approach to adaptive attacks, some recent defense papers do make an effort to handle them.

\citet{struppek_be_2024} reports how NegLS performs against various changes in MIA loss function, learning rate, and initial latent vector sampling. Interestingly, while a raised learning rate and cross-entropy loss are components of our \textit{C-LR} adaptive attack, \citet{struppek_be_2024} did not test these two changes in combination, and so was unable to spot the increased MIA vulnerability we report for NegLS. This is an indication that even seemingly-thorough efforts to search the space of adaptive attacks can occasionally fall short.

\citet{liu2024trapmid} explore specific adaptive attacks where the adversary has different levels of familiarity with (1) the existence of trapdoors and (2) the specific trapdoor signatures. 

\citet{tran2025random} explore one type of adaptive attack where adversaries are aware of the sections of images that are randomly being erased during optimization, which proved not to be effective against MIDRE as a defense.

Overall, adaptive attack coverage in MIA defense work remains limited, raising the risk that defense success is over-estimated.

\subsection{The Manifold Hypothesis and MIA}
\label{app:manifold-related-work}
The manifold hypothesis \citep{fefferman2016testing, bordt2023manifold} argues that real-world high-dimensional data distributions $p$ tend to lie on a relatively low-dimensional manifold $\mathcal{M}$. In the setting of model inversion, we can say that the private dataset $\mathcal{D}_\text{pri}$ has a corresponding low-dimensional manifold $\mathcal{M}_\text{pri}$, and the attack GAN $G$ ranges over a manifold $\mathcal{M}_\text{aux}$, hopefully similar to $\mathcal{M}_\text{pri}$ \citep{peng2026generative}. $\mathcal{M}_\text{pri}$ can be called the \textit{data manifold} and $\mathcal{M}_\text{aux}$ the \textit{generator manifold} \citep{peng2026generative}.

\section{AI Use}
We used generative AI tools to assist with proofreading and polishing both writeup and code. However, we did not use generative AI for research ideation or to originate the hypotheses studied in this paper. All AI-assisted writing was reviewed by the authors, and all AI-assisted code changes were manually reviewed and tested. The authors take responsibility for the text, code, and results in this work.

\section{Dual Use and Ethical Considerations}
This work develops adaptive model inversion attacks. The corresponding ethical risk is that stronger attacks can also improve an adversary's ability to extract information about private training data. However, the primary purpose of our work is defensive evaluation: we show that current best-practice MIA evaluations can substantially underestimate the degree to which trained models expose recognizable information about their training data, including biometric information such as facial geometry. We also characterize a tradeoff between this privacy leakage and another Trustworthy-ML objective, adversarial robustness. Better measurement of both vulnerabilities and their tradeoffs provides a stronger basis for designing and evaluating privacy defenses. It also provides practitioners with a better understanding of the risks to which they expose the humans represented in their datasets.

Nonetheless, we deliberately restrict our experiments to widely studied public datasets, public model architectures, and experimental settings already common in recent MIA research.  Our experiments reproduce standard high-resolution MIA settings and modify the attack and evaluation procedures used to measure leakage.

\section{Deferred Alignment Results}

\begin{figure}[H]
    \centering
    \includegraphics[width=0.45\linewidth]{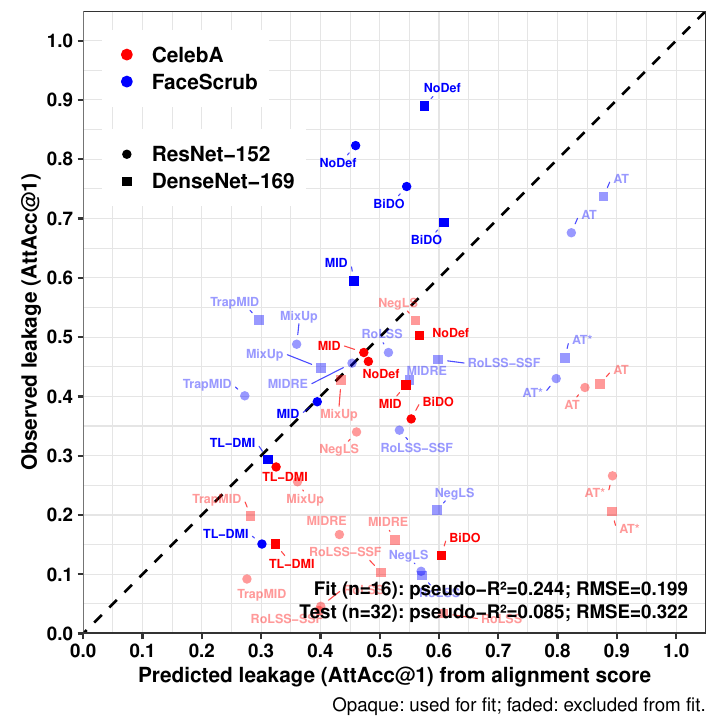}
    \caption{Regression predicting leakage from the alignment score. The fitted alignment coefficient is positive (6.16), consistent with greater alignment being associated with greater leakage, although held-out predictive fit is weak.}
    \label{fig:regression-alignment}
\end{figure}

\textbf{Figure \ref{fig:regression-alignment}} reports 15-class alignment scores regressed against MIA AttAcc@1.

\section{Deferred Regression Results}

\begin{figure}[H]
    \centering
    \begin{subfigure}{0.45\linewidth}
        \centering
        \includegraphics[width=\linewidth]{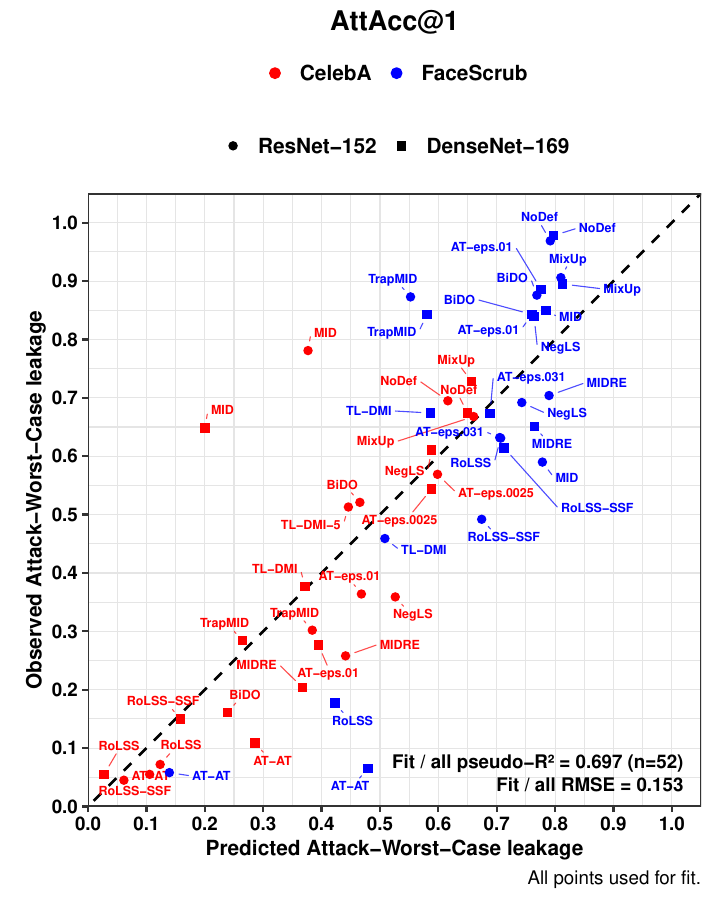}
        \caption{AttAcc@1}
        \label{fig:adaptive_all_defenses}
    \end{subfigure}
    \hfill
    \begin{subfigure}{0.45\linewidth}
        \centering
        \includegraphics[width=\linewidth]{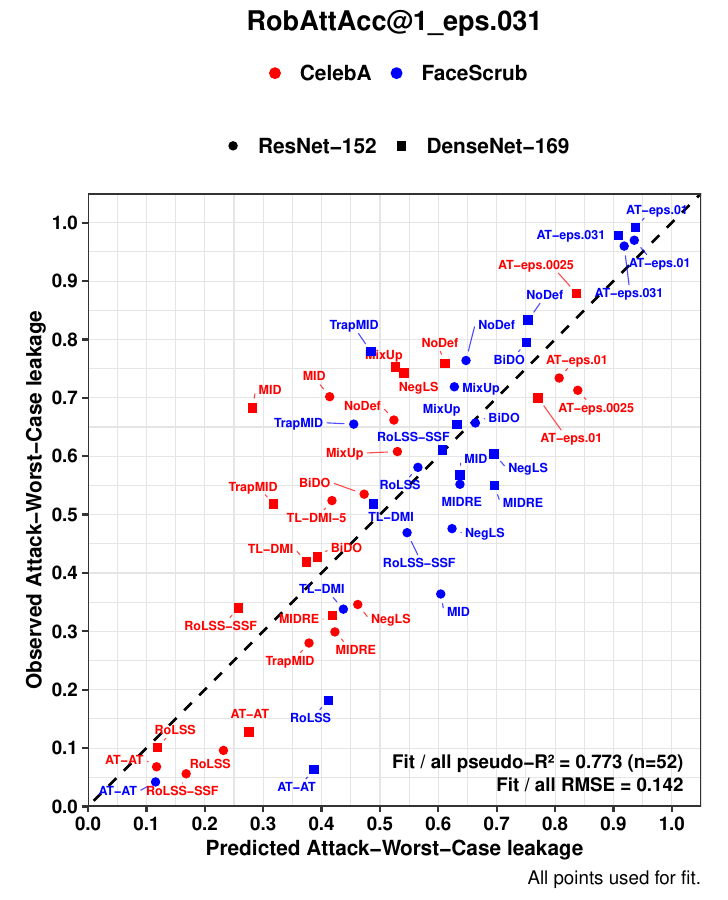}
        \caption{Robust-AttAcc@1}
        \label{fig:adaptive_all_defenses_robust}
    \end{subfigure}

    \caption{Regressions fit on all data points using the worst-case adaptive attack.}
    \label{fig:adaptive_all_defenses_combined}
\end{figure}

\textbf{Figure \ref{fig:adaptive_all_defenses_combined}} showcases AutoAttack regressions against AttAcc@1 and Robust AttAcc@1 fit over all 52 training regimes, including AT-AT \citep{torp2026reducing}.

\section{Deferred Model Inversion Results}
\label{deferred-model-inversion-results}
\textbf{Table \ref{tab:alignment}} reports mean 15-class alignment scores across (non-adaptive) PPA runs.

\textbf{Table \ref{tab:resnet152-ifgmi-worst-case}} reports an initial attempt applying \textit{N-LR} and \textit{C-LR} adaptive attacks to the IF-GMI attack \citep{qiu_closer_2024} on ResNet-152 models trained on FaceScrub. While these specific adaptive attacks are less effective overall, we still see a significant leakage increase on the NegLS defense. We hypothesize that further improvement may be obtained by (for example) trying multiple learning rate increases, rather than fixing the increase at an arbitrary 10x, though we leave such optimizations to future work. 

\textbf{Table \ref{tab:resnet152-ifgmi-standard-robust}} shows that robust evaluation models still reveal significant additional leakage for AT models under IF-GMI. \textbf{Table \ref{tab:resnet152-facescrub-full-adaptive-ifgmi}} details full IF-GMI adaptive attack results.

\textbf{Table \ref{tab:autoacc}} reports AutoAttack worst-case robustness measurements across four epsilon levels; three epsilon values were used in all of the robustness regressions (0.0025, 0.0005, 1e-5), where $\epsilon=0.031$ is shown to indicate its lack of descriptiveness in the MIA setting.

\textbf{Tables \ref{tab:resnet152-facescrub-celeba-full-adaptive-ppa} and \ref{tab:densenet169-facescrub-celeba-full-adaptive-ppa}} showcase complete \textit{N-LR} and \textit{C-LR} adaptive attack results on PPA.

\newpage

\begin{table}[h]
\centering
\begin{compacttable}
\caption{Alignment scores of the target models calculated with vanilla PPA.}
\label{tab:alignment}
\setlength{\tabcolsep}{3pt}
\begin{adjustbox}{max width=\linewidth}
\begin{tabular}{@{}l*{4}{c}@{}}
\toprule
 & \multicolumn{2}{c}{\textbf{FaceScrub}} & \multicolumn{2}{c}{\textbf{CelebA}} \\
\cmidrule(lr){2-3}\cmidrule(lr){4-5}
\textbf{Model} & ResNet-152 & DenseNet-169 & ResNet-152 & DenseNet-169 \\
\midrule
NoDef & 0.144 & 0.220 & 0.158 & 0.214 \\
\midrule
MID & 0.101 & 0.142 & 0.153 & 0.199 \\
BiDO & 0.200 & 0.242 & 0.205 & 0.239 \\
TL\mbox{-}DMI & 0.034 & 0.042 & 0.052 & 0.051 \\
RoLSS & 0.180 & 0.217 & 0.105 & 0.241 \\
RoLSS\mbox{-}SSF & 0.192 & 0.235 & 0.103 & 0.172 \\
MIDRE & 0.140 & 0.203 & 0.126 & 0.187 \\
Trap\mbox{-}MID & 0.011 & 0.030 & 0.014 & 0.019 \\
NegLS & 0.216 & 0.234 & 0.145 & 0.210 \\
\midrule
AT\mbox{-}AT & 0.021 & 0.031 & 0.018 & 0.019 \\
\midrule
MixUp & 0.077 & 0.105 & 0.078 & 0.128 \\
AT & 0.420 & 0.490 & 0.447 & 0.481 \\
AT* & 0.393 & 0.408 & 0.514 & 0.512 \\
\bottomrule
\end{tabular}
\end{adjustbox}
\end{compacttable}
\end{table}

\begin{table}[t]
\newcommand{\ifgmibar}{\rule[0.4ex]{0.80em}{0.23em}}
\centering
\small
\setlength{\tabcolsep}{3pt}
\renewcommand{\arraystretch}{1.05}
\caption{ResNet-152 adaptive IF-GMI attack results on FaceScrub. Each attack
metric shows standard IF-GMI $\rightarrow$ worst-case adaptive IF-GMI over
IF-GMI, IF-GMI N-LR, and IF-GMI C-LR: maximum AttAcc@1 and minimum L2-Face,
computed independently. Small parenthesized symbols identify the attack
attaining each worst-case value (all symbols are shown for ties).
TestAcc is from standard IF-GMI. AT and AT* use $\epsilon=0.01$ and
$0.031$, respectively.}
\label{tab:resnet152-ifgmi-worst-case}
\begin{adjustbox}{max width=\linewidth}
\begin{tabular}{@{}lccc@{}}
\toprule
& \multicolumn{3}{c}{\textbf{FaceScrub}} \\
\cmidrule(lr){2-4}
\textbf{Model} & \textbf{TestAcc} $\uparrow$ & \textbf{AttAcc@1} $\downarrow$ & \textbf{L2-Face} $\uparrow$ \\
\midrule
NoDef & 0.943 & $0.976 \rightarrow 0.976\,{\scriptstyle (\text{\ifgmibar})}$ & $0.724 \rightarrow 0.724\,{\scriptstyle (\text{\ifgmibar})}$ \\
\midrule
MID & 0.950 & $0.380 \rightarrow 0.380\,{\scriptstyle (\text{\ifgmibar})}$ & $1.013 \rightarrow 1.013\,{\scriptstyle (\text{\ifgmibar})}$ \\
BiDO & 0.920 & $0.935 \rightarrow 0.935\,{\scriptstyle (\text{\ifgmibar})}$ & $0.797 \rightarrow 0.797\,{\scriptstyle (\text{\ifgmibar})}$ \\
TL\mbox{-}DMI & 0.891 & $0.383 \rightarrow 0.383\,{\scriptstyle (\text{\ifgmibar})}$ & $0.962 \rightarrow 0.962\,{\scriptstyle (\text{\ifgmibar})}$ \\
RoLSS & 0.886 & $0.512 \rightarrow 0.546\,{\scriptstyle (\bigstar)}$ & $0.908 \rightarrow 0.908\,{\scriptstyle (\text{\ifgmibar})}$ \\
RoLSS\mbox{-}SSF & 0.869 & $0.340 \rightarrow 0.399\,{\scriptstyle (\bigstar)}$ & $0.979 \rightarrow 0.953\,{\scriptstyle (\bigstar)}$ \\
MIDRE & 0.941 & $0.774 \rightarrow 0.774\,{\scriptstyle (\text{\ifgmibar})}$ & $0.780 \rightarrow 0.780\,{\scriptstyle (\text{\ifgmibar})}$ \\
Trap\mbox{-}MID & 0.926 & $0.633 \rightarrow 0.633\,{\scriptstyle (\text{\ifgmibar})}$ & $0.982 \rightarrow 0.982\,{\scriptstyle (\text{\ifgmibar})}$ \\
NegLS & 0.911 & $0.087 \rightarrow 0.532\,{\scriptstyle (\bigstar)}$ & $1.269 \rightarrow 0.884\,{\scriptstyle (\bigstar)}$ \\
AT\mbox{-}AT & 0.941 & $0.071 \rightarrow 0.071\,{\scriptstyle (\text{\ifgmibar})}$ & $1.220 \rightarrow 1.220\,{\scriptstyle (\text{\ifgmibar})}$ \\
\midrule
MixUp & 0.971 & $0.761 \rightarrow 0.761\,{\scriptstyle (\text{\ifgmibar})}$ & $0.832 \rightarrow 0.832\,{\scriptstyle (\text{\ifgmibar})}$ \\
AT & 0.928 & $0.899 \rightarrow 0.899\,{\scriptstyle (\text{\ifgmibar})}$ & $0.730 \rightarrow 0.730\,{\scriptstyle (\text{\ifgmibar})}$ \\
AT* & 0.891 & $0.677 \rightarrow 0.677\,{\scriptstyle (\text{\ifgmibar})}$ & $0.788 \rightarrow 0.788\,{\scriptstyle (\text{\ifgmibar})}$ \\
\bottomrule
\end{tabular}
\end{adjustbox}
\par\smallskip
{\footnotesize IF-GMI (baseline): $\text{\ifgmibar}$\qquad
IF-GMI N-LR: $\bigstar$\qquad IF-GMI C-LR: $\blacktriangle$}
\end{table}

\begin{table}[t]
\centering
\small
\setlength{\tabcolsep}{3pt}
\renewcommand{\arraystretch}{1.05}
\caption{ResNet-152 standard and robust IF-GMI attack accuracy on FaceScrub.
Each pair shows AttAcc@$k$ $\rightarrow$ Robust-AttAcc@$k$ for $k\in\{1,5\}$ from
the same standard IF-GMI run. Bold values mark changes $>$$0.1$.}
\label{tab:resnet152-ifgmi-standard-robust}
\begin{adjustbox}{max width=\linewidth}
\begin{tabular}{@{}lccc@{}}
\toprule
& \multicolumn{3}{c}{\textbf{FaceScrub}} \\
\cmidrule(lr){2-4}
\textbf{Model} & \textbf{TestAcc} $\uparrow$ &
\textbf{AttAcc@1} $\downarrow$ &
\textbf{AttAcc@5} $\downarrow$ \\
\midrule
NoDef & 0.943 & \textbf{0.976} $\rightarrow$ \textbf{0.754} & 0.998 $\rightarrow$ 0.922 \\
\midrule
MID & 0.950 & \textbf{0.380} $\rightarrow$ \textbf{0.192} & \textbf{0.654} $\rightarrow$ \textbf{0.437} \\
BiDO & 0.920 & \textbf{0.935} $\rightarrow$ \textbf{0.696} & 0.988 $\rightarrow$ 0.889 \\
TL\mbox{-}DMI & 0.891 & \textbf{0.383} $\rightarrow$ \textbf{0.264} & \textbf{0.667} $\rightarrow$ \textbf{0.522} \\
RoLSS & 0.886 & 0.512 $\rightarrow$ 0.492 & 0.770 $\rightarrow$ 0.784 \\
RoLSS\mbox{-}SSF & 0.869 & 0.340 $\rightarrow$ 0.339 & 0.628 $\rightarrow$ 0.644 \\
MIDRE & 0.941 & \textbf{0.774} $\rightarrow$ \textbf{0.574} & \textbf{0.949} $\rightarrow$ \textbf{0.820} \\
Trap\mbox{-}MID & 0.926 & \textbf{0.633} $\rightarrow$ \textbf{0.427} & \textbf{0.762} $\rightarrow$ \textbf{0.643} \\
NegLS & 0.911 & 0.087 $\rightarrow$ 0.053 & 0.248 $\rightarrow$ 0.177 \\
\midrule
AT\mbox{-}AT & 0.941 & 0.071 $\rightarrow$ 0.049 & 0.215 $\rightarrow$ 0.156 \\
\midrule
MixUp & 0.971 & \textbf{0.761} $\rightarrow$ \textbf{0.545} & \textbf{0.870} $\rightarrow$ \textbf{0.733} \\
AT & 0.928 & 0.899 $\rightarrow$ 0.986 & 0.985 $\rightarrow$ 0.999 \\
AT* & 0.891 & \textbf{0.677} $\rightarrow$ \textbf{0.975} & \textbf{0.880} $\rightarrow$ \textbf{0.996} \\
\bottomrule
\end{tabular}
\end{adjustbox}
\end{table}

\begin{table}[t]
\centering
\begin{compacttable}
\caption{Full adaptive IF\mbox{-}GMI attack results on \textbf{ResNet-152}
trained on FaceScrub.}
\label{tab:resnet152-facescrub-full-adaptive-ifgmi}
\setlength{\tabcolsep}{3pt}
\begin{adjustbox}{max width=\linewidth, max totalheight=\textheight}
\begin{tabular}{@{}lll*{6}{c}@{}}
\toprule
\textbf{Data} &
\textbf{Model} &
\textbf{Attack} &
\textbf{TestAcc $\uparrow$} &
\textbf{AttAcc@1 $\downarrow$} &
\textbf{Robust-AttAcc@1 $\downarrow$} &
\textbf{AttAcc@5 $\downarrow$} &
\textbf{Robust-AttAcc@5 $\downarrow$} &
\textbf{L2-Face $\uparrow$} \\
\midrule
\multirow{39}{*}{\rotatebox[origin=c]{90}{\textbf{FaceScrub}}} & \multirow{3}{*}{NoDef} & IF\mbox{-}GMI & 0.943 & 0.976 & 0.754 & 0.998 & 0.922 & 0.724 \\
 &  & IF\mbox{-}GMI N\mbox{-}LR & 0.943 & 0.908 & 0.665 & 0.988 & 0.895 & 0.799 \\
 &  & IF\mbox{-}GMI C\mbox{-}LR & 0.943 & 0.781 & 0.450 & 0.944 & 0.737 & 0.888 \\
\cmidrule{2-9}
 & \multirow{3}{*}{MID} & IF\mbox{-}GMI & 0.950 & 0.380 & 0.192 & 0.654 & 0.437 & 1.013 \\
 &  & IF\mbox{-}GMI N\mbox{-}LR & 0.950 & 0.294 & 0.172 & 0.519 & 0.373 & 1.105 \\
 &  & IF\mbox{-}GMI C\mbox{-}LR & 0.950 & 0.299 & 0.180 & 0.509 & 0.360 & 1.107 \\
\cmidrule{2-9}
 & \multirow{3}{*}{BiDO} & IF\mbox{-}GMI & 0.920 & 0.935 & 0.696 & 0.988 & 0.889 & 0.797 \\
 &  & IF\mbox{-}GMI N\mbox{-}LR & 0.920 & 0.797 & 0.568 & 0.955 & 0.844 & 0.892 \\
 &  & IF\mbox{-}GMI C\mbox{-}LR & 0.920 & 0.457 & 0.252 & 0.758 & 0.534 & 1.010 \\
\cmidrule{2-9}
 & \multirow{3}{*}{TL\mbox{-}DMI} & IF\mbox{-}GMI & 0.891 & 0.383 & 0.264 & 0.667 & 0.522 & 0.962 \\
 &  & IF\mbox{-}GMI N\mbox{-}LR & 0.891 & 0.290 & 0.214 & 0.552 & 0.450 & 1.024 \\
 &  & IF\mbox{-}GMI C\mbox{-}LR & 0.891 & 0.231 & 0.169 & 0.474 & 0.391 & 1.051 \\
\cmidrule{2-9}
 & \multirow{3}{*}{RoLSS} & IF\mbox{-}GMI & 0.886 & 0.512 & 0.492 & 0.770 & 0.784 & 0.908 \\
 &  & IF\mbox{-}GMI N\mbox{-}LR & 0.886 & 0.546 & 0.486 & 0.831 & 0.807 & 0.920 \\
 &  & IF\mbox{-}GMI C\mbox{-}LR & 0.886 & 0.494 & 0.446 & 0.744 & 0.734 & 0.935 \\
\cmidrule{2-9}
 & \multirow{3}{*}{RoLSS\mbox{-}SSF} & IF\mbox{-}GMI & 0.869 & 0.340 & 0.339 & 0.628 & 0.644 & 0.979 \\
 &  & IF\mbox{-}GMI N\mbox{-}LR & 0.869 & 0.399 & 0.366 & 0.696 & 0.697 & 0.953 \\
 &  & IF\mbox{-}GMI C\mbox{-}LR & 0.869 & 0.350 & 0.358 & 0.619 & 0.644 & 0.988 \\
\cmidrule{2-9}
 & \multirow{3}{*}{MIDRE} & IF\mbox{-}GMI & 0.941 & 0.774 & 0.574 & 0.949 & 0.820 & 0.780 \\
 &  & IF\mbox{-}GMI N\mbox{-}LR & 0.941 & 0.600 & 0.438 & 0.868 & 0.737 & 0.870 \\
 &  & IF\mbox{-}GMI C\mbox{-}LR & 0.941 & 0.308 & 0.239 & 0.614 & 0.490 & 0.973 \\
\cmidrule{2-9}
 & \multirow{3}{*}{Trap\mbox{-}MID} & IF\mbox{-}GMI & 0.926 & 0.633 & 0.427 & 0.762 & 0.643 & 0.982 \\
 &  & IF\mbox{-}GMI N\mbox{-}LR & 0.926 & 0.594 & 0.408 & 0.746 & 0.617 & 1.002 \\
 &  & IF\mbox{-}GMI C\mbox{-}LR & 0.926 & 0.568 & 0.390 & 0.723 & 0.599 & 1.021 \\
\cmidrule{2-9}
 & \multirow{3}{*}{NegLS} & IF\mbox{-}GMI & 0.911 & 0.087 & 0.053 & 0.248 & 0.177 & 1.269 \\
 &  & IF\mbox{-}GMI N\mbox{-}LR & 0.911 & 0.532 & 0.342 & 0.840 & 0.663 & 0.884 \\
 &  & IF\mbox{-}GMI C\mbox{-}LR & 0.911 & 0.297 & 0.177 & 0.606 & 0.427 & 1.006 \\
\cmidrule{2-9}
 & \multirow{3}{*}{AT\mbox{-}AT} & IF\mbox{-}GMI & 0.941 & 0.071 & 0.049 & 0.215 & 0.156 & 1.220 \\
 &  & IF\mbox{-}GMI N\mbox{-}LR & 0.941 & 0.025 & 0.019 & 0.088 & 0.070 & 1.367 \\
 &  & IF\mbox{-}GMI C\mbox{-}LR & 0.941 & 0.023 & 0.017 & 0.083 & 0.068 & 1.369 \\
\cmidrule{2-9}
 & \multirow{3}{*}{MixUp} & IF\mbox{-}GMI & 0.971 & 0.761 & 0.545 & 0.870 & 0.733 & 0.832 \\
 &  & IF\mbox{-}GMI N\mbox{-}LR & 0.971 & 0.720 & 0.525 & 0.849 & 0.718 & 0.887 \\
 &  & IF\mbox{-}GMI C\mbox{-}LR & 0.971 & 0.736 & 0.547 & 0.858 & 0.730 & 0.874 \\
\cmidrule{2-9}
 & \multirow{3}{*}{AT} & IF\mbox{-}GMI & 0.928 & 0.899 & 0.986 & 0.985 & 0.999 & 0.730 \\
 &  & IF\mbox{-}GMI N\mbox{-}LR & 0.928 & 0.779 & 0.923 & 0.950 & 0.996 & 0.858 \\
 &  & IF\mbox{-}GMI C\mbox{-}LR & 0.928 & 0.658 & 0.944 & 0.872 & 0.991 & 0.874 \\
\cmidrule{2-9}
 & \multirow{3}{*}{AT*} & IF\mbox{-}GMI & 0.891 & 0.677 & 0.975 & 0.880 & 0.996 & 0.788 \\
 &  & IF\mbox{-}GMI N\mbox{-}LR & 0.891 & 0.563 & 0.910 & 0.830 & 0.987 & 0.915 \\
 &  & IF\mbox{-}GMI C\mbox{-}LR & 0.891 & 0.496 & 0.932 & 0.740 & 0.987 & 0.934 \\
\bottomrule
\end{tabular}
\end{adjustbox}
\end{compacttable}
\end{table}

\begin{table}[t]
\centering
\begin{compacttable}
\caption{Test accuracy and AutoAttack robust accuracy of the target models.}
\label{tab:autoacc}
\setlength{\tabcolsep}{3pt}
\begin{adjustbox}{max width=\linewidth, max totalheight=\textheight}
\begin{tabular}{@{}lll*{5}{c}@{}}
\toprule
\textbf{Data} &
\textbf{Arch.} &
\textbf{Model} &
\textbf{TestAcc $\uparrow$} &
\textbf{AutoAcc.031} &
\textbf{AutoAcc.0025} &
\textbf{AutoAcc.0005} &
\textbf{AutoAcc1e-5} \\
\midrule
\multirow{26}{*}{\rotatebox[origin=c]{90}{\textbf{FaceScrub}}} & \multirow{13}{*}{\rotatebox[origin=c]{90}{ResNet-152}} & NoDef & 0.943 & 0.000 & 0.054 & 0.853 & 0.939 \\
\cmidrule{3-8}
 &  & MID & 0.950 & 0.000 & 0.006 & 0.770 & 0.928 \\
 &  & BiDO & 0.920 & 0.000 & 0.103 & 0.841 & 0.927 \\
 &  & TL\mbox{-}DMI & 0.891 & 0.000 & 0.000 & 0.000 & 0.836 \\
 &  & RoLSS & 0.886 & 0.000 & 0.009 & 0.744 & 0.882 \\
 &  & RoLSS\mbox{-}SSF & 0.869 & 0.000 & 0.013 & 0.721 & 0.857 \\
 &  & MIDRE & 0.941 & 0.000 & 0.039 & 0.850 & 0.937 \\
 &  & Trap\mbox{-}MID & 0.926 & 0.000 & 0.000 & 0.000 & 0.856 \\
 &  & NegLS & 0.911 & 0.000 & 0.068 & 0.784 & 0.906 \\
\cmidrule{3-8}
 &  & AT\mbox{-}AT & 0.941 & 0.000 & 0.000 & 0.000 & 0.119 \\
\cmidrule{3-8}
 &  & MixUp & 0.971 & 0.000 & 0.000 & 0.789 & 0.964 \\
 &  & AT & 0.928 & 0.002 & 0.891 & 0.920 & 0.926 \\
 &  & AT* & 0.891 & 0.128 & 0.848 & 0.877 & 0.884 \\
\cmidrule{2-8}
 & \multirow{13}{*}{\rotatebox[origin=c]{90}{DenseNet-169}} & NoDef & 0.947 & 0.000 & 0.244 & 0.888 & 0.947 \\
\cmidrule{3-8}
 &  & MID & 0.949 & 0.000 & 0.059 & 0.867 & 0.912 \\
 &  & BiDO & 0.921 & 0.000 & 0.284 & 0.864 & 0.913 \\
 &  & TL\mbox{-}DMI & 0.921 & 0.000 & 0.000 & 0.005 & 0.910 \\
 &  & RoLSS & 0.732 & 0.000 & 0.002 & 0.514 & 0.727 \\
 &  & RoLSS\mbox{-}SSF & 0.886 & 0.000 & 0.070 & 0.780 & 0.884 \\
 &  & MIDRE & 0.929 & 0.000 & 0.177 & 0.842 & 0.912 \\
 &  & Trap\mbox{-}MID & 0.920 & 0.000 & 0.000 & 0.000 & 0.904 \\
 &  & NegLS & 0.924 & 0.000 & 0.170 & 0.836 & 0.919 \\
\cmidrule{3-8}
 &  & AT\mbox{-}AT & 0.933 & 0.000 & 0.000 & 0.000 & 0.744 \\
\cmidrule{3-8}
 &  & MixUp & 0.967 & 0.000 & 0.000 & 0.812 & 0.968 \\
 &  & AT & 0.943 & 0.000 & 0.891 & 0.928 & 0.937 \\
 &  & AT* & 0.892 & 0.036 & 0.827 & 0.855 & 0.861 \\
\midrule
\multirow{26}{*}{\rotatebox[origin=c]{90}{\textbf{CelebA}}} & \multirow{13}{*}{\rotatebox[origin=c]{90}{ResNet-152}} & NoDef & 0.845 & 0.000 & 0.036 & 0.678 & 0.813 \\
\cmidrule{3-8}
 &  & MID & 0.802 & 0.004 & 0.139 & 0.542 & 0.574 \\
 &  & BiDO & 0.747 & 0.000 & 0.062 & 0.590 & 0.748 \\
 &  & TL\mbox{-}DMI & 0.814 & 0.000 & 0.000 & 0.119 & 0.799 \\
 &  & RoLSS & 0.537 & 0.000 & 0.000 & 0.086 & 0.530 \\
 &  & RoLSS\mbox{-}SSF & 0.428 & 0.000 & 0.000 & 0.042 & 0.406 \\
 &  & MIDRE & 0.756 & 0.000 & 0.021 & 0.545 & 0.715 \\
 &  & Trap\mbox{-}MID & 0.806 & 0.000 & 0.000 & 0.000 & 0.760 \\
 &  & NegLS & 0.805 & 0.000 & 0.013 & 0.559 & 0.771 \\
\cmidrule{3-8}
 &  & AT\mbox{-}AT & 0.819 & 0.000 & 0.000 & 0.000 & 0.155 \\
\cmidrule{3-8}
 &  & MixUp & 0.899 & 0.000 & 0.000 & 0.463 & 0.888 \\
 &  & AT & 0.818 & 0.000 & 0.638 & 0.797 & 0.819 \\
 &  & AT* & 0.748 & 0.001 & 0.658 & 0.725 & 0.744 \\
\cmidrule{2-8}
 & \multirow{13}{*}{\rotatebox[origin=c]{90}{DenseNet-169}} & NoDef & 0.856 & 0.000 & 0.142 & 0.750 & 0.833 \\
\cmidrule{3-8}
 &  & MID & 0.730 & 0.011 & 0.129 & 0.405 & 0.396 \\
 &  & BiDO & 0.611 & 0.000 & 0.143 & 0.484 & 0.594 \\
 &  & TL\mbox{-}DMI & 0.788 & 0.000 & 0.000 & 0.044 & 0.748 \\
 &  & RoLSS & 0.275 & 0.000 & 0.001 & 0.122 & 0.267 \\
 &  & RoLSS\mbox{-}SSF & 0.557 & 0.000 & 0.001 & 0.289 & 0.538 \\
 &  & MIDRE & 0.696 & 0.000 & 0.061 & 0.535 & 0.685 \\
 &  & Trap\mbox{-}MID & 0.711 & 0.000 & 0.000 & 0.021 & 0.673 \\
 &  & NegLS & 0.820 & 0.000 & 0.077 & 0.670 & 0.808 \\
\cmidrule{3-8}
 &  & AT\mbox{-}AT & 0.836 & 0.000 & 0.000 & 0.000 & 0.567 \\
\cmidrule{3-8}
 &  & MixUp & 0.871 & 0.000 & 0.000 & 0.642 & 0.854 \\
 &  & AT & 0.827 & 0.000 & 0.651 & 0.774 & 0.802 \\
 &  & AT* & 0.701 & 0.001 & 0.629 & 0.689 & 0.705 \\
\bottomrule
\end{tabular}
\end{adjustbox}
\end{compacttable}
\end{table}

\begin{table}[t]
\centering
\begin{compacttable}
\caption{Full adaptive PPA attack results on \textbf{ResNet-152}
trained on FaceScrub and CelebA.}
\label{tab:resnet152-facescrub-celeba-full-adaptive-ppa}
\setlength{\tabcolsep}{3pt}
\begin{adjustbox}{max width=\linewidth, max totalheight=\textheight}
\begin{tabular}{@{}lll*{6}{c}@{}}
\toprule
\textbf{Data} &
\textbf{Model} &
\textbf{Attack} &
\textbf{TestAcc $\uparrow$} &
\textbf{AttAcc@1 $\downarrow$} &
\textbf{Robust-AttAcc@1 $\downarrow$} &
\textbf{AttAcc@5 $\downarrow$} &
\textbf{Robust-AttAcc@5 $\downarrow$} &
\textbf{L2-Face $\uparrow$} \\
\midrule
\multirow{39}{*}{\rotatebox[origin=c]{90}{\textbf{FaceScrub}}} & \multirow{3}{*}{NoDef} & PPA & 0.943 & 0.823 & 0.463 & 0.961 & 0.761 & 0.806 \\
 &  & PPA N\mbox{-}LR & 0.943 & 0.969 & 0.764 & 0.999 & 0.945 & 0.712 \\
 &  & PPA C\mbox{-}LR & 0.943 & 0.861 & 0.578 & 0.979 & 0.829 & 0.805 \\
\cmidrule{2-9}
 & \multirow{3}{*}{MID} & PPA & 0.950 & 0.391 & 0.219 & 0.699 & 0.489 & 0.950 \\
 &  & PPA N\mbox{-}LR & 0.950 & 0.563 & 0.358 & 0.828 & 0.657 & 0.878 \\
 &  & PPA C\mbox{-}LR & 0.950 & 0.590 & 0.364 & 0.844 & 0.657 & 0.867 \\
\cmidrule{2-9}
 & \multirow{3}{*}{BiDO} & PPA & 0.920 & 0.754 & 0.494 & 0.937 & 0.781 & 0.832 \\
 &  & PPA N\mbox{-}LR & 0.920 & 0.876 & 0.657 & 0.989 & 0.913 & 0.797 \\
 &  & PPA C\mbox{-}LR & 0.920 & 0.575 & 0.366 & 0.848 & 0.668 & 0.925 \\
\cmidrule{2-9}
 & \multirow{3}{*}{TL\mbox{-}DMI} & PPA & 0.891 & 0.151 & 0.131 & 0.384 & 0.318 & 1.088 \\
 &  & PPA N\mbox{-}LR & 0.891 & 0.459 & 0.338 & 0.750 & 0.630 & 0.954 \\
 &  & PPA C\mbox{-}LR & 0.891 & 0.367 & 0.289 & 0.665 & 0.574 & 0.986 \\
\cmidrule{2-9}
 & \multirow{3}{*}{RoLSS} & PPA & 0.886 & 0.474 & 0.470 & 0.760 & 0.777 & 0.858 \\
 &  & PPA N\mbox{-}LR & 0.886 & 0.631 & 0.581 & 0.894 & 0.880 & 0.832 \\
 &  & PPA C\mbox{-}LR & 0.886 & 0.584 & 0.560 & 0.836 & 0.850 & 0.850 \\
\cmidrule{2-9}
 & \multirow{3}{*}{RoLSS\mbox{-}SSF} & PPA & 0.869 & 0.343 & 0.350 & 0.635 & 0.653 & 0.918 \\
 &  & PPA N\mbox{-}LR & 0.869 & 0.492 & 0.469 & 0.786 & 0.788 & 0.894 \\
 &  & PPA C\mbox{-}LR & 0.869 & 0.442 & 0.439 & 0.704 & 0.737 & 0.910 \\
\cmidrule{2-9}
 & \multirow{3}{*}{MIDRE} & PPA & 0.941 & 0.456 & 0.343 & 0.761 & 0.624 & 0.891 \\
 &  & PPA N\mbox{-}LR & 0.941 & 0.704 & 0.552 & 0.928 & 0.829 & 0.815 \\
 &  & PPA C\mbox{-}LR & 0.941 & 0.421 & 0.333 & 0.725 & 0.621 & 0.923 \\
\cmidrule{2-9}
 & \multirow{3}{*}{Trap\mbox{-}MID} & PPA & 0.926 & 0.401 & 0.232 & 0.676 & 0.481 & 0.980 \\
 &  & PPA N\mbox{-}LR & 0.926 & 0.873 & 0.655 & 0.970 & 0.891 & 0.777 \\
 &  & PPA C\mbox{-}LR & 0.926 & 0.851 & 0.626 & 0.967 & 0.869 & 0.800 \\
\cmidrule{2-9}
 & \multirow{3}{*}{NegLS} & PPA & 0.911 & 0.105 & 0.069 & 0.277 & 0.200 & 1.217 \\
 &  & PPA N\mbox{-}LR & 0.911 & 0.692 & 0.476 & 0.945 & 0.806 & 0.796 \\
 &  & PPA C\mbox{-}LR & 0.911 & 0.467 & 0.291 & 0.797 & 0.608 & 0.907 \\
\cmidrule{2-9}
 & \multirow{3}{*}{AT\mbox{-}AT} & PPA & 0.941 & 0.058 & 0.042 & 0.190 & 0.145 & 1.225 \\
 &  & PPA N\mbox{-}LR & 0.941 & 0.049 & 0.037 & 0.163 & 0.134 & 1.264 \\
 &  & PPA C\mbox{-}LR & 0.941 & 0.046 & 0.036 & 0.168 & 0.128 & 1.279 \\
\cmidrule{2-9}
 & \multirow{3}{*}{MixUp} & PPA & 0.971 & 0.488 & 0.331 & 0.727 & 0.588 & 0.932 \\
 &  & PPA N\mbox{-}LR & 0.971 & 0.871 & 0.681 & 0.957 & 0.857 & 0.758 \\
 &  & PPA C\mbox{-}LR & 0.971 & 0.906 & 0.719 & 0.970 & 0.883 & 0.725 \\
\cmidrule{2-9}
 & \multirow{3}{*}{AT} & PPA & 0.928 & 0.676 & 0.934 & 0.896 & 0.995 & 0.758 \\
 &  & PPA N\mbox{-}LR & 0.928 & 0.842 & 0.965 & 0.972 & 0.998 & 0.766 \\
 &  & PPA C\mbox{-}LR & 0.928 & 0.717 & 0.970 & 0.908 & 0.998 & 0.789 \\
\cmidrule{2-9}
 & \multirow{3}{*}{AT*} & PPA & 0.891 & 0.430 & 0.901 & 0.715 & 0.982 & 0.849 \\
 &  & PPA N\mbox{-}LR & 0.891 & 0.632 & 0.942 & 0.879 & 0.997 & 0.829 \\
 &  & PPA C\mbox{-}LR & 0.891 & 0.545 & 0.960 & 0.797 & 0.994 & 0.858 \\
\midrule
\multirow{39}{*}{\rotatebox[origin=c]{90}{\textbf{CelebA}}} & \multirow{3}{*}{NoDef} & PPA & 0.845 & 0.459 & 0.518 & 0.750 & 0.786 & 0.852 \\
 &  & PPA N\mbox{-}LR & 0.845 & 0.695 & 0.662 & 0.912 & 0.897 & 0.838 \\
 &  & PPA C\mbox{-}LR & 0.845 & 0.459 & 0.483 & 0.718 & 0.746 & 0.892 \\
\cmidrule{2-9}
 & \multirow{3}{*}{MID} & PPA & 0.802 & 0.474 & 0.441 & 0.725 & 0.688 & 0.821 \\
 &  & PPA N\mbox{-}LR & 0.802 & 0.781 & 0.702 & 0.939 & 0.891 & 0.709 \\
 &  & PPA C\mbox{-}LR & 0.802 & 0.761 & 0.670 & 0.935 & 0.880 & 0.714 \\
\cmidrule{2-9}
 & \multirow{3}{*}{BiDO} & PPA & 0.747 & 0.362 & 0.387 & 0.629 & 0.653 & 0.923 \\
 &  & PPA N\mbox{-}LR & 0.747 & 0.521 & 0.535 & 0.784 & 0.796 & 0.876 \\
 &  & PPA C\mbox{-}LR & 0.747 & 0.294 & 0.335 & 0.565 & 0.613 & 0.982 \\
\cmidrule{2-9}
 & \multirow{3}{*}{TL\mbox{-}DMI} & PPA & 0.814 & 0.281 & 0.311 & 0.550 & 0.591 & 0.929 \\
 &  & PPA N\mbox{-}LR & 0.814 & 0.513 & 0.524 & 0.786 & 0.815 & 0.863 \\
 &  & PPA C\mbox{-}LR & 0.814 & 0.338 & 0.402 & 0.617 & 0.667 & 0.930 \\
\cmidrule{2-9}
 & \multirow{3}{*}{RoLSS} & PPA & 0.537 & 0.046 & 0.066 & 0.145 & 0.182 & 1.147 \\
 &  & PPA N\mbox{-}LR & 0.537 & 0.072 & 0.095 & 0.192 & 0.231 & 1.124 \\
 &  & PPA C\mbox{-}LR & 0.537 & 0.066 & 0.096 & 0.190 & 0.242 & 1.117 \\
\cmidrule{2-9}
 & \multirow{3}{*}{RoLSS\mbox{-}SSF} & PPA & 0.428 & 0.039 & 0.030 & 0.118 & 0.097 & 1.183 \\
 &  & PPA N\mbox{-}LR & 0.428 & 0.043 & 0.054 & 0.136 & 0.155 & 1.187 \\
 &  & PPA C\mbox{-}LR & 0.428 & 0.045 & 0.056 & 0.147 & 0.154 & 1.186 \\
\cmidrule{2-9}
 & \multirow{3}{*}{MIDRE} & PPA & 0.756 & 0.167 & 0.223 & 0.377 & 0.461 & 0.987 \\
 &  & PPA N\mbox{-}LR & 0.756 & 0.258 & 0.299 & 0.514 & 0.581 & 0.951 \\
 &  & PPA C\mbox{-}LR & 0.756 & 0.165 & 0.215 & 0.383 & 0.462 & 1.019 \\
\cmidrule{2-9}
 & \multirow{3}{*}{Trap\mbox{-}MID} & PPA & 0.806 & 0.092 & 0.097 & 0.193 & 0.195 & 1.358 \\
 &  & PPA N\mbox{-}LR & 0.806 & 0.302 & 0.280 & 0.482 & 0.465 & 1.150 \\
 &  & PPA C\mbox{-}LR & 0.806 & 0.245 & 0.246 & 0.418 & 0.415 & 1.165 \\
\cmidrule{2-9}
 & \multirow{3}{*}{NegLS} & PPA & 0.805 & 0.340 & 0.333 & 0.623 & 0.639 & 0.908 \\
 &  & PPA N\mbox{-}LR & 0.805 & 0.359 & 0.346 & 0.686 & 0.675 & 0.897 \\
 &  & PPA C\mbox{-}LR & 0.805 & 0.250 & 0.262 & 0.538 & 0.553 & 0.935 \\
\cmidrule{2-9}
 & \multirow{3}{*}{AT\mbox{-}AT} & PPA & 0.819 & 0.049 & 0.059 & 0.138 & 0.151 & 1.245 \\
 &  & PPA N\mbox{-}LR & 0.819 & 0.055 & 0.068 & 0.149 & 0.177 & 1.261 \\
 &  & PPA C\mbox{-}LR & 0.819 & 0.048 & 0.064 & 0.142 & 0.178 & 1.263 \\
\cmidrule{2-9}
 & \multirow{3}{*}{MixUp} & PPA & 0.899 & 0.256 & 0.294 & 0.483 & 0.511 & 0.965 \\
 &  & PPA N\mbox{-}LR & 0.899 & 0.629 & 0.550 & 0.834 & 0.779 & 0.822 \\
 &  & PPA C\mbox{-}LR & 0.899 & 0.668 & 0.608 & 0.850 & 0.817 & 0.792 \\
\cmidrule{2-9}
 & \multirow{3}{*}{AT} & PPA & 0.818 & 0.415 & 0.713 & 0.693 & 0.912 & 0.822 \\
 &  & PPA N\mbox{-}LR & 0.818 & 0.569 & 0.671 & 0.795 & 0.903 & 0.863 \\
 &  & PPA C\mbox{-}LR & 0.818 & 0.416 & 0.700 & 0.685 & 0.886 & 0.866 \\
\cmidrule{2-9}
 & \multirow{3}{*}{AT*} & PPA & 0.748 & 0.266 & 0.734 & 0.527 & 0.929 & 0.887 \\
 &  & PPA N\mbox{-}LR & 0.748 & 0.364 & 0.703 & 0.626 & 0.927 & 0.911 \\
 &  & PPA C\mbox{-}LR & 0.748 & 0.271 & 0.721 & 0.507 & 0.907 & 0.937 \\
\bottomrule
\end{tabular}
\end{adjustbox}
\end{compacttable}
\end{table}

\begin{table}[t]
\centering
\begin{compacttable}
\caption{Full adaptive PPA attack results on \textbf{DenseNet-169}
trained on FaceScrub and CelebA.}
\label{tab:densenet169-facescrub-celeba-full-adaptive-ppa}
\setlength{\tabcolsep}{3pt}
\begin{adjustbox}{max width=\linewidth, max totalheight=\textheight}
\begin{tabular}{@{}lll*{6}{c}@{}}
\toprule
\textbf{Data} &
\textbf{Model} &
\textbf{Attack} &
\textbf{TestAcc $\uparrow$} &
\textbf{AttAcc@1 $\downarrow$} &
\textbf{Robust-AttAcc@1 $\downarrow$} &
\textbf{AttAcc@5 $\downarrow$} &
\textbf{Robust-AttAcc@5 $\downarrow$} &
\textbf{L2-Face $\uparrow$} \\
\midrule
\multirow{39}{*}{\rotatebox[origin=c]{90}{\textbf{FaceScrub}}} & \multirow{3}{*}{NoDef} & PPA & 0.947 & 0.890 & 0.641 & 0.980 & 0.889 & 0.763 \\
 &  & PPA N\mbox{-}LR & 0.947 & 0.978 & 0.833 & 0.999 & 0.974 & 0.703 \\
 &  & PPA C\mbox{-}LR & 0.947 & 0.828 & 0.589 & 0.964 & 0.860 & 0.835 \\
\cmidrule{2-9}
 & \multirow{3}{*}{MID} & PPA & 0.949 & 0.595 & 0.306 & 0.835 & 0.575 & 0.869 \\
 &  & PPA N\mbox{-}LR & 0.949 & 0.826 & 0.552 & 0.967 & 0.825 & 0.755 \\
 &  & PPA C\mbox{-}LR & 0.949 & 0.850 & 0.568 & 0.970 & 0.837 & 0.743 \\
\cmidrule{2-9}
 & \multirow{3}{*}{BiDO} & PPA & 0.921 & 0.693 & 0.614 & 0.891 & 0.866 & 0.831 \\
 &  & PPA N\mbox{-}LR & 0.921 & 0.843 & 0.795 & 0.967 & 0.949 & 0.788 \\
 &  & PPA C\mbox{-}LR & 0.921 & 0.643 & 0.614 & 0.874 & 0.861 & 0.874 \\
\cmidrule{2-9}
 & \multirow{3}{*}{TL\mbox{-}DMI} & PPA & 0.921 & 0.294 & 0.191 & 0.575 & 0.443 & 0.996 \\
 &  & PPA N\mbox{-}LR & 0.921 & 0.674 & 0.518 & 0.911 & 0.806 & 0.851 \\
 &  & PPA C\mbox{-}LR & 0.921 & 0.484 & 0.359 & 0.764 & 0.651 & 0.928 \\
\cmidrule{2-9}
 & \multirow{3}{*}{RoLSS} & PPA & 0.732 & 0.098 & 0.108 & 0.267 & 0.267 & 1.116 \\
 &  & PPA N\mbox{-}LR & 0.732 & 0.177 & 0.182 & 0.424 & 0.423 & 1.050 \\
 &  & PPA C\mbox{-}LR & 0.732 & 0.169 & 0.180 & 0.412 & 0.408 & 1.058 \\
\cmidrule{2-9}
 & \multirow{3}{*}{RoLSS\mbox{-}SSF} & PPA & 0.886 & 0.462 & 0.415 & 0.761 & 0.738 & 0.882 \\
 &  & PPA N\mbox{-}LR & 0.886 & 0.614 & 0.611 & 0.851 & 0.883 & 0.838 \\
 &  & PPA C\mbox{-}LR & 0.886 & 0.598 & 0.606 & 0.840 & 0.862 & 0.851 \\
\cmidrule{2-9}
 & \multirow{3}{*}{MIDRE} & PPA & 0.929 & 0.428 & 0.377 & 0.734 & 0.674 & 0.899 \\
 &  & PPA N\mbox{-}LR & 0.929 & 0.651 & 0.550 & 0.906 & 0.836 & 0.839 \\
 &  & PPA C\mbox{-}LR & 0.929 & 0.339 & 0.330 & 0.642 & 0.621 & 0.965 \\
\cmidrule{2-9}
 & \multirow{3}{*}{Trap\mbox{-}MID} & PPA & 0.920 & 0.529 & 0.442 & 0.787 & 0.720 & 0.893 \\
 &  & PPA N\mbox{-}LR & 0.920 & 0.843 & 0.779 & 0.959 & 0.949 & 0.772 \\
 &  & PPA C\mbox{-}LR & 0.920 & 0.711 & 0.678 & 0.910 & 0.888 & 0.842 \\
\cmidrule{2-9}
 & \multirow{3}{*}{NegLS} & PPA & 0.924 & 0.208 & 0.138 & 0.450 & 0.323 & 1.107 \\
 &  & PPA N\mbox{-}LR & 0.924 & 0.839 & 0.604 & 0.971 & 0.895 & 0.712 \\
 &  & PPA C\mbox{-}LR & 0.924 & 0.478 & 0.323 & 0.798 & 0.652 & 0.899 \\
\cmidrule{2-9}
 & \multirow{3}{*}{AT\mbox{-}AT} & PPA & 0.933 & 0.058 & 0.063 & 0.190 & 0.164 & 1.194 \\
 &  & PPA N\mbox{-}LR & 0.933 & 0.056 & 0.057 & 0.203 & 0.176 & 1.240 \\
 &  & PPA C\mbox{-}LR & 0.933 & 0.065 & 0.060 & 0.205 & 0.173 & 1.236 \\
\cmidrule{2-9}
 & \multirow{3}{*}{MixUp} & PPA & 0.967 & 0.448 & 0.293 & 0.704 & 0.525 & 0.960 \\
 &  & PPA N\mbox{-}LR & 0.967 & 0.865 & 0.601 & 0.958 & 0.790 & 0.801 \\
 &  & PPA C\mbox{-}LR & 0.967 & 0.895 & 0.654 & 0.965 & 0.824 & 0.768 \\
\cmidrule{2-9}
 & \multirow{3}{*}{AT} & PPA & 0.943 & 0.737 & 0.959 & 0.920 & 0.996 & 0.734 \\
 &  & PPA N\mbox{-}LR & 0.943 & 0.886 & 0.992 & 0.983 & 0.999 & 0.709 \\
 &  & PPA C\mbox{-}LR & 0.943 & 0.581 & 0.907 & 0.818 & 0.985 & 0.857 \\
\cmidrule{2-9}
 & \multirow{3}{*}{AT*} & PPA & 0.892 & 0.465 & 0.919 & 0.734 & 0.986 & 0.839 \\
 &  & PPA N\mbox{-}LR & 0.892 & 0.673 & 0.954 & 0.897 & 0.997 & 0.785 \\
 &  & PPA C\mbox{-}LR & 0.892 & 0.640 & 0.978 & 0.861 & 0.998 & 0.803 \\
\midrule
\multirow{39}{*}{\rotatebox[origin=c]{90}{\textbf{CelebA}}} & \multirow{3}{*}{NoDef} & PPA & 0.856 & 0.503 & 0.681 & 0.788 & 0.897 & 0.800 \\
 &  & PPA N\mbox{-}LR & 0.856 & 0.674 & 0.759 & 0.886 & 0.936 & 0.810 \\
 &  & PPA C\mbox{-}LR & 0.856 & 0.474 & 0.616 & 0.741 & 0.840 & 0.874 \\
\cmidrule{2-9}
 & \multirow{3}{*}{MID} & PPA & 0.730 & 0.419 & 0.458 & 0.675 & 0.718 & 0.836 \\
 &  & PPA N\mbox{-}LR & 0.730 & 0.648 & 0.683 & 0.853 & 0.872 & 0.753 \\
 &  & PPA C\mbox{-}LR & 0.730 & 0.610 & 0.636 & 0.833 & 0.854 & 0.764 \\
\cmidrule{2-9}
 & \multirow{3}{*}{BiDO} & PPA & 0.611 & 0.132 & 0.339 & 0.313 & 0.607 & 1.021 \\
 &  & PPA N\mbox{-}LR & 0.611 & 0.161 & 0.427 & 0.374 & 0.674 & 1.025 \\
 &  & PPA C\mbox{-}LR & 0.611 & 0.123 & 0.356 & 0.300 & 0.598 & 1.064 \\
\cmidrule{2-9}
 & \multirow{3}{*}{TL\mbox{-}DMI} & PPA & 0.788 & 0.151 & 0.178 & 0.361 & 0.402 & 1.000 \\
 &  & PPA N\mbox{-}LR & 0.788 & 0.377 & 0.419 & 0.654 & 0.703 & 0.904 \\
 &  & PPA C\mbox{-}LR & 0.788 & 0.279 & 0.329 & 0.524 & 0.591 & 0.957 \\
\cmidrule{2-9}
 & \multirow{3}{*}{RoLSS} & PPA & 0.275 & 0.034 & 0.077 & 0.126 & 0.195 & 1.149 \\
 &  & PPA N\mbox{-}LR & 0.275 & 0.048 & 0.101 & 0.166 & 0.242 & 1.143 \\
 &  & PPA C\mbox{-}LR & 0.275 & 0.055 & 0.100 & 0.155 & 0.234 & 1.150 \\
\cmidrule{2-9}
 & \multirow{3}{*}{RoLSS\mbox{-}SSF} & PPA & 0.557 & 0.103 & 0.234 & 0.285 & 0.482 & 1.007 \\
 &  & PPA N\mbox{-}LR & 0.557 & 0.137 & 0.311 & 0.352 & 0.591 & 0.996 \\
 &  & PPA C\mbox{-}LR & 0.557 & 0.150 & 0.340 & 0.358 & 0.590 & 1.008 \\
\cmidrule{2-9}
 & \multirow{3}{*}{MIDRE} & PPA & 0.696 & 0.158 & 0.266 & 0.351 & 0.510 & 0.990 \\
 &  & PPA N\mbox{-}LR & 0.696 & 0.204 & 0.327 & 0.429 & 0.602 & 0.994 \\
 &  & PPA C\mbox{-}LR & 0.696 & 0.160 & 0.273 & 0.352 & 0.508 & 1.020 \\
\cmidrule{2-9}
 & \multirow{3}{*}{Trap\mbox{-}MID} & PPA & 0.711 & 0.198 & 0.401 & 0.443 & 0.661 & 0.957 \\
 &  & PPA N\mbox{-}LR & 0.711 & 0.273 & 0.518 & 0.545 & 0.769 & 0.949 \\
 &  & PPA C\mbox{-}LR & 0.711 & 0.284 & 0.514 & 0.532 & 0.757 & 0.946 \\
\cmidrule{2-9}
 & \multirow{3}{*}{NegLS} & PPA & 0.820 & 0.528 & 0.674 & 0.800 & 0.916 & 0.781 \\
 &  & PPA N\mbox{-}LR & 0.820 & 0.611 & 0.743 & 0.877 & 0.933 & 0.733 \\
 &  & PPA C\mbox{-}LR & 0.820 & 0.297 & 0.396 & 0.604 & 0.703 & 0.882 \\
\cmidrule{2-9}
 & \multirow{3}{*}{AT\mbox{-}AT} & PPA & 0.836 & 0.086 & 0.091 & 0.205 & 0.220 & 1.177 \\
 &  & PPA N\mbox{-}LR & 0.836 & 0.109 & 0.128 & 0.255 & 0.275 & 1.151 \\
 &  & PPA C\mbox{-}LR & 0.836 & 0.108 & 0.128 & 0.264 & 0.285 & 1.152 \\
\cmidrule{2-9}
 & \multirow{3}{*}{MixUp} & PPA & 0.871 & 0.428 & 0.517 & 0.690 & 0.748 & 0.866 \\
 &  & PPA N\mbox{-}LR & 0.871 & 0.695 & 0.694 & 0.873 & 0.870 & 0.790 \\
 &  & PPA C\mbox{-}LR & 0.871 & 0.728 & 0.753 & 0.883 & 0.895 & 0.762 \\
\cmidrule{2-9}
 & \multirow{3}{*}{AT} & PPA & 0.827 & 0.421 & 0.797 & 0.690 & 0.949 & 0.816 \\
 &  & PPA N\mbox{-}LR & 0.827 & 0.544 & 0.879 & 0.809 & 0.975 & 0.792 \\
 &  & PPA C\mbox{-}LR & 0.827 & 0.386 & 0.749 & 0.647 & 0.915 & 0.879 \\
\cmidrule{2-9}
 & \multirow{3}{*}{AT*} & PPA & 0.701 & 0.206 & 0.675 & 0.425 & 0.879 & 0.932 \\
 &  & PPA N\mbox{-}LR & 0.701 & 0.277 & 0.700 & 0.508 & 0.887 & 0.919 \\
 &  & PPA C\mbox{-}LR & 0.701 & 0.183 & 0.639 & 0.389 & 0.847 & 0.987 \\
\bottomrule
\end{tabular}
\end{adjustbox}
\end{compacttable}
\end{table}

\end{document}